\newcounter{treecount}
\newcounter{branchcount}
\newsavebox{\parentbox}
\newsavebox{\treebox}
\newsavebox{\treeboxone}
\newsavebox{\treeboxtwo}
\newsavebox{\treeboxthree}
\newsavebox{\treeboxfour}
\newsavebox{\treeboxfive}
\newsavebox{\treeboxsix}
\newsavebox{\treeboxseven}
\newsavebox{\treeboxeight}
\newsavebox{\treeboxnine}
\newsavebox{\treeboxten}
\newsavebox{\treeboxeleven}
\newsavebox{\treeboxtwelve}
\newsavebox{\treeboxthirteen}
\newsavebox{\treeboxfourteen}
\newsavebox{\treeboxfifteen}
\newsavebox{\treeboxsixteen}
\newsavebox{\treeboxseventeen}
\newsavebox{\treeboxeighteen}
\newsavebox{\treeboxnineteen}
\newsavebox{\treeboxtwenty}
\newlength{\treeoffsetone}
\newlength{\treeoffsettwo}
\newlength{\treeoffsetthree}
\newlength{\treeoffsetfour}
\newlength{\treeoffsetfive}
\newlength{\treeoffsetsix}
\newlength{\treeoffsetseven}
\newlength{\treeoffseteight}
\newlength{\treeoffsetnine}
\newlength{\treeoffsetten}
\newlength{\treeoffseteleven}
\newlength{\treeoffsettwelve}
\newlength{\treeoffsetthirteen}
\newlength{\treeoffsetfourteen}
\newlength{\treeoffsetfifteen}
\newlength{\treeoffsetsixteen}
\newlength{\treeoffsetseventeen}
\newlength{\treeoffseteighteen}
\newlength{\treeoffsetnineteen}
\newlength{\treeoffsettwenty}

\newlength{\treeshiftone}
\newlength{\treeshifttwo}
\newlength{\treeshiftthree}
\newlength{\treeshiftfour}
\newlength{\treeshiftfive}
\newlength{\treeshiftsix}
\newlength{\treeshiftseven}
\newlength{\treeshifteight}
\newlength{\treeshiftnine}
\newlength{\treeshiftten}
\newlength{\treeshifteleven}
\newlength{\treeshifttwelve}
\newlength{\treeshiftthirteen}
\newlength{\treeshiftfourteen}
\newlength{\treeshiftfifteen}
\newlength{\treeshiftsixteen}
\newlength{\treeshiftseventeen}
\newlength{\treeshifteighteen}
\newlength{\treeshiftnineteen}
\newlength{\treeshifttwenty}
\newlength{\treewidthone}
\newlength{\treewidthtwo}
\newlength{\treewidththree}
\newlength{\treewidthfour}
\newlength{\treewidthfive}
\newlength{\treewidthsix}
\newlength{\treewidthseven}
\newlength{\treewidtheight}
\newlength{\treewidthnine}
\newlength{\treewidthten}
\newlength{\treewidtheleven}
\newlength{\treewidthtwelve}
\newlength{\treewidththirteen}
\newlength{\treewidthfourteen}
\newlength{\treewidthfifteen}
\newlength{\treewidthsixteen}
\newlength{\treewidthseventeen}
\newlength{\treewidtheighteen}
\newlength{\treewidthnineteen}
\newlength{\treewidthtwenty}
\newlength{\daughteroffsetone}
\newlength{\daughteroffsettwo}
\newlength{\daughteroffsetthree}
\newlength{\daughteroffsetfour}
\newlength{\branchwidthone}
\newlength{\branchwidthtwo}
\newlength{\branchwidththree}
\newlength{\branchwidthfour}
\newlength{\parentoffset}
\newlength{\treeoffset}
\newlength{\daughteroffset}
\newlength{\branchwidth}
\newlength{\parentwidth}
\newlength{\treewidth}
\newcommand{\ontop}[1]{\begin{tabular}{c}#1\end{tabular}}
\newcommand{\poptree}{%
\ifnum\value{treecount}=0\typeout{QobiTeX warning---Tree stack underflow}\fi%
\addtocounter{treecount}{-1}%
\setlength{\treeoffsettwo}{\treeoffsetthree}%
\setlength{\treeoffsetthree}{\treeoffsetfour}%
\setlength{\treeoffsetfour}{\treeoffsetfive}%
\setlength{\treeoffsetfive}{\treeoffsetsix}%
\setlength{\treeoffsetsix}{\treeoffsetseven}%
\setlength{\treeoffsetseven}{\treeoffseteight}%
\setlength{\treeoffseteight}{\treeoffsetnine}%
\setlength{\treeoffsetnine}{\treeoffsetten}%
\setlength{\treeoffsetten}{\treeoffseteleven}%
\setlength{\treeoffseteleven}{\treeoffsettwelve}%
\setlength{\treeoffsettwelve}{\treeoffsetthirteen}%
\setlength{\treeoffsetthirteen}{\treeoffsetfourteen}%
\setlength{\treeoffsetfourteen}{\treeoffsetfifteen}%
\setlength{\treeoffsetfifteen}{\treeoffsetsixteen}%
\setlength{\treeoffsetsixteen}{\treeoffsetseventeen}%
\setlength{\treeoffsetseventeen}{\treeoffseteighteen}%
\setlength{\treeoffseteighteen}{\treeoffsetnineteen}%
\setlength{\treeoffsetnineteen}{\treeoffsettwenty}%
\setlength{\treeshifttwo}{\treeshiftthree}%
\setlength{\treeshiftthree}{\treeshiftfour}%
\setlength{\treeshiftfour}{\treeshiftfive}%
\setlength{\treeshiftfive}{\treeshiftsix}%
\setlength{\treeshiftsix}{\treeshiftseven}%
\setlength{\treeshiftseven}{\treeshifteight}%
\setlength{\treeshifteight}{\treeshiftnine}%
\setlength{\treeshiftnine}{\treeshiftten}%
\setlength{\treeshiftten}{\treeshifteleven}%
\setlength{\treeshifteleven}{\treeshifttwelve}%
\setlength{\treeshifttwelve}{\treeshiftthirteen}%
\setlength{\treeshiftthirteen}{\treeshiftfourteen}%
\setlength{\treeshiftfourteen}{\treeshiftfifteen}%
\setlength{\treeshiftfifteen}{\treeshiftsixteen}%
\setlength{\treeshiftsixteen}{\treeshiftseventeen}%
\setlength{\treeshiftseventeen}{\treeshifteighteen}%
\setlength{\treeshifteighteen}{\treeshiftnineteen}%
\setlength{\treeshiftnineteen}{\treeshifttwenty}%
\setlength{\treewidthtwo}{\treewidththree}%
\setlength{\treewidththree}{\treewidthfour}%
\setlength{\treewidthfour}{\treewidthfive}%
\setlength{\treewidthfive}{\treewidthsix}%
\setlength{\treewidthsix}{\treewidthseven}%
\setlength{\treewidthseven}{\treewidtheight}%
\setlength{\treewidtheight}{\treewidthnine}%
\setlength{\treewidthnine}{\treewidthten}%
\setlength{\treewidthten}{\treewidtheleven}%
\setlength{\treewidtheleven}{\treewidthtwelve}%
\setlength{\treewidthtwelve}{\treewidththirteen}%
\setlength{\treewidththirteen}{\treewidthfourteen}%
\setlength{\treewidthfourteen}{\treewidthfifteen}%
\setlength{\treewidthfifteen}{\treewidthsixteen}%
\setlength{\treewidthsixteen}{\treewidthseventeen}%
\setlength{\treewidthseventeen}{\treewidtheighteen}%
\setlength{\treewidtheighteen}{\treewidthnineteen}%
\setlength{\treewidthnineteen}{\treewidthtwenty}%
\sbox{\treeboxtwo}{\usebox{\treeboxthree}}%
\sbox{\treeboxthree}{\usebox{\treeboxfour}}%
\sbox{\treeboxfour}{\usebox{\treeboxfive}}%
\sbox{\treeboxfive}{\usebox{\treeboxsix}}%
\sbox{\treeboxsix}{\usebox{\treeboxseven}}%
\sbox{\treeboxseven}{\usebox{\treeboxeight}}%
\sbox{\treeboxeight}{\usebox{\treeboxnine}}%
\sbox{\treeboxnine}{\usebox{\treeboxten}}%
\sbox{\treeboxten}{\usebox{\treeboxeleven}}%
\sbox{\treeboxeleven}{\usebox{\treeboxtwelve}}%
\sbox{\treeboxtwelve}{\usebox{\treeboxthirteen}}%
\sbox{\treeboxthirteen}{\usebox{\treeboxfourteen}}%
\sbox{\treeboxfourteen}{\usebox{\treeboxfifteen}}%
\sbox{\treeboxfifteen}{\usebox{\treeboxsixteen}}%
\sbox{\treeboxsixteen}{\usebox{\treeboxseventeen}}%
\sbox{\treeboxseventeen}{\usebox{\treeboxeighteen}}%
\sbox{\treeboxeighteen}{\usebox{\treeboxnineteen}}%
\sbox{\treeboxnineteen}{\usebox{\treeboxtwenty}}}
\newcommand{\leaf}[1]{%
\ifnum\value{treecount}=20\typeout{QobiTeX warning---Tree stack overflow}\fi%
\addtocounter{treecount}{1}%
\sbox{\treeboxtwenty}{\usebox{\treeboxnineteen}}%
\sbox{\treeboxnineteen}{\usebox{\treeboxeighteen}}%
\sbox{\treeboxeighteen}{\usebox{\treeboxseventeen}}%
\sbox{\treeboxseventeen}{\usebox{\treeboxsixteen}}%
\sbox{\treeboxsixteen}{\usebox{\treeboxfifteen}}%
\sbox{\treeboxfifteen}{\usebox{\treeboxfourteen}}%
\sbox{\treeboxfourteen}{\usebox{\treeboxthirteen}}%
\sbox{\treeboxthirteen}{\usebox{\treeboxtwelve}}%
\sbox{\treeboxtwelve}{\usebox{\treeboxeleven}}%
\sbox{\treeboxeleven}{\usebox{\treeboxten}}%
\sbox{\treeboxten}{\usebox{\treeboxnine}}%
\sbox{\treeboxnine}{\usebox{\treeboxeight}}%
\sbox{\treeboxeight}{\usebox{\treeboxseven}}%
\sbox{\treeboxseven}{\usebox{\treeboxsix}}%
\sbox{\treeboxsix}{\usebox{\treeboxfive}}%
\sbox{\treeboxfive}{\usebox{\treeboxfour}}%
\sbox{\treeboxfour}{\usebox{\treeboxthree}}%
\sbox{\treeboxthree}{\usebox{\treeboxtwo}}%
\sbox{\treeboxtwo}{\usebox{\treeboxone}}%
\sbox{\treeboxone}{\ontop{#1}}%
\sbox{\treeboxone}{\raisebox{-\ht\treeboxone}{\usebox{\treeboxone}}}%
\setlength{\treeoffsettwenty}{\treeoffsetnineteen}%
\setlength{\treeoffsetnineteen}{\treeoffseteighteen}%
\setlength{\treeoffseteighteen}{\treeoffsetseventeen}%
\setlength{\treeoffsetseventeen}{\treeoffsetsixteen}%
\setlength{\treeoffsetsixteen}{\treeoffsetfifteen}%
\setlength{\treeoffsetfifteen}{\treeoffsetfourteen}%
\setlength{\treeoffsetfourteen}{\treeoffsetthirteen}%
\setlength{\treeoffsetthirteen}{\treeoffsettwelve}%
\setlength{\treeoffsettwelve}{\treeoffseteleven}%
\setlength{\treeoffseteleven}{\treeoffsetten}%
\setlength{\treeoffsetten}{\treeoffsetnine}%
\setlength{\treeoffsetnine}{\treeoffseteight}%
\setlength{\treeoffseteight}{\treeoffsetseven}%
\setlength{\treeoffsetseven}{\treeoffsetsix}%
\setlength{\treeoffsetsix}{\treeoffsetfive}%
\setlength{\treeoffsetfive}{\treeoffsetfour}%
\setlength{\treeoffsetfour}{\treeoffsetthree}%
\setlength{\treeoffsetthree}{\treeoffsettwo}%
\setlength{\treeoffsettwo}{\treeoffsetone}%
\setlength{\treeoffsetone}{0.5\wd\treeboxone}%
\setlength{\treeshifttwenty}{\treeshiftnineteen}%
\setlength{\treeshiftnineteen}{\treeshifteighteen}%
\setlength{\treeshifteighteen}{\treeshiftseventeen}%
\setlength{\treeshiftseventeen}{\treeshiftsixteen}%
\setlength{\treeshiftsixteen}{\treeshiftfifteen}%
\setlength{\treeshiftfifteen}{\treeshiftfourteen}%
\setlength{\treeshiftfourteen}{\treeshiftthirteen}%
\setlength{\treeshiftthirteen}{\treeshifttwelve}%
\setlength{\treeshifttwelve}{\treeshifteleven}%
\setlength{\treeshifteleven}{\treeshiftten}%
\setlength{\treeshiftten}{\treeshiftnine}%
\setlength{\treeshiftnine}{\treeshifteight}%
\setlength{\treeshifteight}{\treeshiftseven}%
\setlength{\treeshiftseven}{\treeshiftsix}%
\setlength{\treeshiftsix}{\treeshiftfive}%
\setlength{\treeshiftfive}{\treeshiftfour}%
\setlength{\treeshiftfour}{\treeshiftthree}%
\setlength{\treeshiftthree}{\treeshifttwo}%
\setlength{\treeshifttwo}{\treeshiftone}%
\setlength{\treeshiftone}{0pt}%
\setlength{\treewidthtwenty}{\treewidthnineteen}%
\setlength{\treewidthnineteen}{\treewidtheighteen}%
\setlength{\treewidtheighteen}{\treewidthseventeen}%
\setlength{\treewidthseventeen}{\treewidthsixteen}%
\setlength{\treewidthsixteen}{\treewidthfifteen}%
\setlength{\treewidthfifteen}{\treewidthfourteen}%
\setlength{\treewidthfourteen}{\treewidththirteen}%
\setlength{\treewidththirteen}{\treewidthtwelve}%
\setlength{\treewidthtwelve}{\treewidtheleven}%
\setlength{\treewidtheleven}{\treewidthten}%
\setlength{\treewidthten}{\treewidthnine}%
\setlength{\treewidthnine}{\treewidtheight}%
\setlength{\treewidtheight}{\treewidthseven}%
\setlength{\treewidthseven}{\treewidthsix}%
\setlength{\treewidthsix}{\treewidthfive}%
\setlength{\treewidthfive}{\treewidthfour}%
\setlength{\treewidthfour}{\treewidththree}%
\setlength{\treewidththree}{\treewidthtwo}%
\setlength{\treewidthtwo}{\treewidthone}%
\setlength{\treewidthone}{\wd\treeboxone}}
\newcommand{\branch}[2]{%
\setcounter{branchcount}{#1}%
\ifnum\value{branchcount}=1\sbox{\parentbox}{\ontop{#2}}%
\setlength{\parentoffset}{\treeoffsetone}%
\addtolength{\parentoffset}{-0.5\wd\parentbox}%
\setlength{\daughteroffset}{0in}%
\ifdim\parentoffset<0in%
\setlength{\daughteroffset}{-\parentoffset}%
\setlength{\parentoffset}{0in}\fi%
\setlength{\parentwidth}{\parentoffset}%
\addtolength{\parentwidth}{\wd\parentbox}%
\setlength{\treeoffset}{\daughteroffset}%
\addtolength{\treeoffset}{\treeoffsetone}%
\setlength{\treewidth}{\wd\treeboxone}%
\addtolength{\treewidth}{\daughteroffset}%
\ifdim\treewidth<\parentwidth\setlength{\treewidth}{\parentwidth}\fi%
\sbox{\treebox}{\begin{minipage}{\treewidth}%
\begin{flushleft}%
\hspace*{\parentoffset}\usebox{\parentbox}\\
{\setlength{\unitlength}{2ex}%
\hspace*{\treeoffset}\begin{picture}(0,1)%
\put(0,0){\line(0,1){1}}%
\end{picture}}\\
\vspace{-\baselineskip}
\hspace*{\daughteroffset}%
\raisebox{-\ht\treeboxone}{\usebox{\treeboxone}}%
\end{flushleft}%
\end{minipage}}%
\setlength{\treeoffsetone}{\parentoffset}%
\addtolength{\treeoffsetone}{0.5\wd\parentbox}%
\setlength{\treeshiftone}{0pt}%
\setlength{\treewidthone}{\treewidth}%
\sbox{\treeboxone}{\usebox{\treebox}}%
\else\ifnum\value{branchcount}=2\sbox{\parentbox}{\ontop{#2}}%
\setlength{\branchwidthone}{\treewidthtwo}%
\addtolength{\branchwidthone}{\treeoffsetone}%
\addtolength{\branchwidthone}{-\treeshiftone}%
\addtolength{\branchwidthone}{-\treeoffsettwo}%
\setlength{\branchwidth}{\branchwidthone}%
\setlength{\daughteroffsetone}{\branchwidth}%
\addtolength{\daughteroffsetone}{-\branchwidthone}%
\addtolength{\daughteroffsetone}{-\treeshiftone}%
\setlength{\parentoffset}{-0.5\wd\parentbox}%
\addtolength{\parentoffset}{\treeoffsettwo}%
\addtolength{\parentoffset}{0.5\branchwidth}%
\setlength{\daughteroffset}{0in}%
\ifdim\parentoffset<0in%
\setlength{\daughteroffset}{-\parentoffset}%
\setlength{\parentoffset}{0in}\fi%
\setlength{\parentwidth}{\parentoffset}%
\addtolength{\parentwidth}{\wd\parentbox}%
\setlength{\treeoffset}{\daughteroffset}%
\addtolength{\treeoffset}{\treeoffsettwo}%
\setlength{\treewidth}{\wd\treeboxone}%
\addtolength{\treewidth}{\daughteroffsetone}%
\addtolength{\treewidth}{\treewidthtwo}%
\addtolength{\treewidth}{\daughteroffset}%
\ifdim\treewidth<\parentwidth\setlength{\treewidth}{\parentwidth}\fi%
\sbox{\treebox}{\begin{minipage}{\treewidth}%
\begin{flushleft}%
\hspace*{\parentoffset}\usebox{\parentbox}\\
{\setlength{\unitlength}{0.5\branchwidth}%
\hspace*{\treeoffset}\begin{picture}(2,0.5)%
\put(0,0){\line(2,1){1}}%
\put(2,0){\line(-2,1){1}}%
\end{picture}}\\
\vspace{-\baselineskip}
\hspace*{\daughteroffset}%
\makebox[\treewidthtwo][l]%
{\raisebox{-\ht\treeboxtwo}{\usebox{\treeboxtwo}}}%
\hspace*{\daughteroffsetone}%
\raisebox{-\ht\treeboxone}{\usebox{\treeboxone}}%
\end{flushleft}%
\end{minipage}}%
\setlength{\treeoffsetone}{\parentoffset}%
\addtolength{\treeoffsetone}{0.5\wd\parentbox}%
\setlength{\treeshiftone}{0pt}%
\setlength{\treewidthone}{\treewidth}%
\sbox{\treeboxone}{\usebox{\treebox}}\poptree%
\else\ifnum\value{branchcount}=3\sbox{\parentbox}{\ontop{#2}}%
\setlength{\branchwidthone}{\treewidthtwo}%
\addtolength{\branchwidthone}{\treeoffsetone}%
\addtolength{\branchwidthone}{-\treeshiftone}%
\addtolength{\branchwidthone}{-\treeoffsettwo}%
\setlength{\branchwidthtwo}{\treewidththree}%
\addtolength{\branchwidthtwo}{\treeoffsettwo}%
\addtolength{\branchwidthtwo}{-\treeshifttwo}%
\addtolength{\branchwidthtwo}{-\treeoffsetthree}%
\setlength{\branchwidth}{\branchwidthone}%
\ifdim\branchwidthtwo>\branchwidth%
\setlength{\branchwidth}{\branchwidthtwo}\fi%
\setlength{\daughteroffsetone}{\branchwidth}%
\addtolength{\daughteroffsetone}{-\branchwidthone}%
\addtolength{\daughteroffsetone}{-\treeshiftone}%
\setlength{\daughteroffsettwo}{\branchwidth}%
\addtolength{\daughteroffsettwo}{-\branchwidthtwo}%
\addtolength{\daughteroffsettwo}{-\treeshifttwo}%
\setlength{\parentoffset}{-0.5\wd\parentbox}%
\addtolength{\parentoffset}{\treeoffsetthree}%
\addtolength{\parentoffset}{\branchwidth}%
\setlength{\daughteroffset}{0in}%
\ifdim\parentoffset<0in%
\setlength{\daughteroffset}{-\parentoffset}%
\setlength{\parentoffset}{0in}\fi%
\setlength{\parentwidth}{\parentoffset}%
\addtolength{\parentwidth}{\wd\parentbox}%
\setlength{\treeoffset}{\daughteroffset}%
\addtolength{\treeoffset}{\treeoffsetthree}%
\setlength{\treewidth}{\wd\treeboxone}%
\addtolength{\treewidth}{\daughteroffsetone}%
\addtolength{\treewidth}{\treewidthtwo}%
\addtolength{\treewidth}{\daughteroffsettwo}%
\addtolength{\treewidth}{\treewidththree}%
\addtolength{\treewidth}{\daughteroffset}%
\ifdim\treewidth<\parentwidth\setlength{\treewidth}{\parentwidth}\fi%
\sbox{\treebox}{\begin{minipage}{\treewidth}%
\begin{flushleft}%
\hspace*{\parentoffset}\usebox{\parentbox}\\
{\setlength{\unitlength}{0.5\branchwidth}%
\hspace*{\treeoffset}\begin{picture}(4,1)%
\put(0,0){\line(2,1){2}}%
\put(2,0){\line(0,1){1}}%
\put(4,0){\line(-2,1){2}}%
\end{picture}}\\
\vspace{-\baselineskip}
\hspace*{\daughteroffset}%
\makebox[\treewidththree][l]%
{\raisebox{-\ht\treeboxthree}{\usebox{\treeboxthree}}}%
\hspace*{\daughteroffsettwo}%
\makebox[\treewidthtwo][l]%
{\raisebox{-\ht\treeboxtwo}{\usebox{\treeboxtwo}}}%
\hspace*{\daughteroffsetone}%
\raisebox{-\ht\treeboxone}{\usebox{\treeboxone}}%
\end{flushleft}%
\end{minipage}}%
\setlength{\treeoffsetone}{\parentoffset}%
\addtolength{\treeoffsetone}{0.5\wd\parentbox}%
\setlength{\treeshiftone}{0pt}%
\setlength{\treewidthone}{\treewidth}%
\sbox{\treeboxone}{\usebox{\treebox}}\poptree\poptree%
\else\ifnum\value{branchcount}=4\sbox{\parentbox}{\ontop{#2}}%
\setlength{\branchwidthone}{\treewidthtwo}%
\addtolength{\branchwidthone}{\treeoffsetone}%
\addtolength{\branchwidthone}{-\treeshiftone}%
\addtolength{\branchwidthone}{-\treeoffsettwo}%
\setlength{\branchwidthtwo}{\treewidththree}%
\addtolength{\branchwidthtwo}{\treeoffsettwo}%
\addtolength{\branchwidthtwo}{-\treeshifttwo}%
\addtolength{\branchwidthtwo}{-\treeoffsetthree}%
\setlength{\branchwidththree}{\treewidthfour}%
\addtolength{\branchwidththree}{\treeoffsetthree}%
\addtolength{\branchwidththree}{-\treeshiftthree}%
\addtolength{\branchwidththree}{-\treeoffsetfour}%
\setlength{\branchwidth}{\branchwidthone}%
\ifdim\branchwidthtwo>\branchwidth%
\setlength{\branchwidth}{\branchwidthtwo}\fi%
\ifdim\branchwidththree>\branchwidth%
\setlength{\branchwidth}{\branchwidththree}\fi%
\setlength{\daughteroffsetone}{\branchwidth}%
\addtolength{\daughteroffsetone}{-\branchwidthone}%
\addtolength{\daughteroffsetone}{-\treeshiftone}%
\setlength{\daughteroffsettwo}{\branchwidth}%
\addtolength{\daughteroffsettwo}{-\branchwidthtwo}%
\addtolength{\daughteroffsettwo}{-\treeshifttwo}%
\setlength{\daughteroffsetthree}{\branchwidth}%
\addtolength{\daughteroffsetthree}{-\branchwidththree}%
\addtolength{\daughteroffsetthree}{-\treeshiftthree}%
\setlength{\parentoffset}{-0.5\wd\parentbox}%
\addtolength{\parentoffset}{\treeoffsetfour}%
\addtolength{\parentoffset}{1.5\branchwidth}%
\setlength{\daughteroffset}{0in}%
\ifdim\parentoffset<0in%
\setlength{\daughteroffset}{-\parentoffset}%
\setlength{\parentoffset}{0in}\fi%
\setlength{\parentwidth}{\parentoffset}%
\addtolength{\parentwidth}{\wd\parentbox}%
\setlength{\treeoffset}{\daughteroffset}%
\addtolength{\treeoffset}{\treeoffsetfour}%
\setlength{\treewidth}{\wd\treeboxone}%
\addtolength{\treewidth}{\daughteroffsetone}%
\addtolength{\treewidth}{\treewidthtwo}%
\addtolength{\treewidth}{\daughteroffsettwo}%
\addtolength{\treewidth}{\treewidththree}%
\addtolength{\treewidth}{\daughteroffsetthree}%
\addtolength{\treewidth}{\treewidthfour}%
\addtolength{\treewidth}{\daughteroffset}%
\ifdim\treewidth<\parentwidth\setlength{\treewidth}{\parentwidth}\fi%
\sbox{\treebox}{\begin{minipage}{\treewidth}%
\begin{flushleft}%
\hspace*{\parentoffset}\usebox{\parentbox}\\
{\setlength{\unitlength}{0.5\branchwidth}%
\hspace*{\treeoffset}\begin{picture}(6,1)%
\put(0,0){\line(3,1){3}}%
\put(2,0){\line(1,1){1}}%
\put(4,0){\line(-1,1){1}}%
\put(6,0){\line(-3,1){3}}%
\end{picture}}\\
\vspace{-\baselineskip}
\hspace*{\daughteroffset}%
\makebox[\treewidthfour][l]%
{\raisebox{-\ht\treeboxfour}{\usebox{\treeboxfour}}}%
\hspace*{\daughteroffsetthree}%
\makebox[\treewidththree][l]%
{\raisebox{-\ht\treeboxthree}{\usebox{\treeboxthree}}}%
\hspace*{\daughteroffsettwo}%
\makebox[\treewidthtwo][l]%
{\raisebox{-\ht\treeboxtwo}{\usebox{\treeboxtwo}}}%
\hspace*{\daughteroffsetone}%
\raisebox{-\ht\treeboxone}{\usebox{\treeboxone}}%
\end{flushleft}%
\end{minipage}}%
\setlength{\treeoffsetone}{\parentoffset}%
\addtolength{\treeoffsetone}{0.5\wd\parentbox}%
\setlength{\treeshiftone}{0pt}%
\setlength{\treewidthone}{\treewidth}%
\sbox{\treeboxone}{\usebox{\treebox}}\poptree\poptree\poptree%
\else\ifnum\value{branchcount}=5\sbox{\parentbox}{\ontop{#2}}%
\setlength{\branchwidthone}{\treewidthtwo}%
\addtolength{\branchwidthone}{\treeoffsetone}%
\addtolength{\branchwidthone}{-\treeshiftone}%
\addtolength{\branchwidthone}{-\treeoffsettwo}%
\setlength{\branchwidthtwo}{\treewidththree}%
\addtolength{\branchwidthtwo}{\treeoffsettwo}%
\addtolength{\branchwidthtwo}{-\treeshifttwo}%
\addtolength{\branchwidthtwo}{-\treeoffsetthree}%
\setlength{\branchwidththree}{\treewidthfour}%
\addtolength{\branchwidththree}{\treeoffsetthree}%
\addtolength{\branchwidththree}{-\treeshiftthree}%
\addtolength{\branchwidththree}{-\treeoffsetfour}%
\setlength{\branchwidthfour}{\treewidthfive}%
\addtolength{\branchwidthfour}{\treeoffsetfour}%
\addtolength{\branchwidthfour}{-\treeshiftfour}%
\addtolength{\branchwidthfour}{-\treeoffsetfive}%
\setlength{\branchwidth}{\branchwidthone}%
\ifdim\branchwidthtwo>\branchwidth%
\setlength{\branchwidth}{\branchwidthtwo}\fi%
\ifdim\branchwidththree>\branchwidth%
\setlength{\branchwidth}{\branchwidththree}\fi%
\ifdim\branchwidthfour>\branchwidth%
\setlength{\branchwidth}{\branchwidthfour}\fi%
\setlength{\daughteroffsetone}{\branchwidth}%
\addtolength{\daughteroffsetone}{-\branchwidthone}%
\addtolength{\daughteroffsetone}{-\treeshiftone}%
\setlength{\daughteroffsettwo}{\branchwidth}%
\addtolength{\daughteroffsettwo}{-\branchwidthtwo}%
\addtolength{\daughteroffsettwo}{-\treeshifttwo}%
\setlength{\daughteroffsetthree}{\branchwidth}%
\addtolength{\daughteroffsetthree}{-\branchwidththree}%
\addtolength{\daughteroffsetthree}{-\treeshiftthree}%
\setlength{\daughteroffsetfour}{\branchwidth}%
\addtolength{\daughteroffsetfour}{-\branchwidthfour}%
\addtolength{\daughteroffsetfour}{-\treeshiftfour}%
\setlength{\parentoffset}{-0.5\wd\parentbox}%
\addtolength{\parentoffset}{\treeoffsetfive}%
\addtolength{\parentoffset}{2\branchwidth}%
\setlength{\daughteroffset}{0in}%
\ifdim\parentoffset<0in%
\setlength{\daughteroffset}{-\parentoffset}%
\setlength{\parentoffset}{0in}\fi%
\setlength{\parentwidth}{\parentoffset}%
\addtolength{\parentwidth}{\wd\parentbox}%
\setlength{\treeoffset}{\daughteroffset}%
\addtolength{\treeoffset}{\treeoffsetfive}%
\setlength{\treewidth}{\wd\treeboxone}%
\addtolength{\treewidth}{\daughteroffsetone}%
\addtolength{\treewidth}{\treewidthtwo}%
\addtolength{\treewidth}{\daughteroffsettwo}%
\addtolength{\treewidth}{\treewidththree}%
\addtolength{\treewidth}{\daughteroffsetthree}%
\addtolength{\treewidth}{\treewidthfour}%
\addtolength{\treewidth}{\daughteroffsetfour}%
\addtolength{\treewidth}{\treewidthfive}%
\addtolength{\treewidth}{\daughteroffset}%
\ifdim\treewidth<\parentwidth\setlength{\treewidth}{\parentwidth}\fi%
\sbox{\treebox}{\begin{minipage}{\treewidth}%
\begin{flushleft}%
\hspace*{\parentoffset}\usebox{\parentbox}\\
{\setlength{\unitlength}{0.5\branchwidth}%
\hspace*{\treeoffset}\begin{picture}(8,1)%
\put(0,0){\line(4,1){4}}%
\put(2,0){\line(2,1){2}}%
\put(4,0){\line(0,1){1}}%
\put(6,0){\line(-2,1){2}}%
\put(8,0){\line(-4,1){4}}%
\end{picture}}\\
\vspace{-\baselineskip}
\hspace*{\daughteroffset}%
\makebox[\treewidthfive][l]%
{\raisebox{-\ht\treeboxfour}{\usebox{\treeboxfive}}}%
\hspace*{\daughteroffsetfour}%
\makebox[\treewidthfour][l]%
{\raisebox{-\ht\treeboxfour}{\usebox{\treeboxfour}}}%
\hspace*{\daughteroffsetthree}%
\makebox[\treewidththree][l]%
{\raisebox{-\ht\treeboxthree}{\usebox{\treeboxthree}}}%
\hspace*{\daughteroffsettwo}%
\makebox[\treewidthtwo][l]%
{\raisebox{-\ht\treeboxtwo}{\usebox{\treeboxtwo}}}%
\hspace*{\daughteroffsetone}%
\raisebox{-\ht\treeboxone}{\usebox{\treeboxone}}%
\end{flushleft}%
\end{minipage}}%
\setlength{\treeoffsetone}{\parentoffset}%
\addtolength{\treeoffsetone}{0.5\wd\parentbox}%
\setlength{\treeshiftone}{0pt}%
\setlength{\treewidthone}{\treewidth}%
\sbox{\treeboxone}{\usebox{\treebox}}\poptree\poptree\poptree\poptree%
\else\typeout{QobiTeX warning--- Can't handle #1 branching}\fi\fi\fi\fi\fi}
\newcommand{\faketreewidth}[1]{%
\sbox{\parentbox}{\ontop{#1}}%
\setlength{\treewidthone}{0.5\wd\parentbox}%
\addtolength{\treewidthone}{\treeoffsetone}%
\setlength{\treeshiftone}{\treeoffsetone}%
\addtolength{\treeshiftone}{-0.5\wd\parentbox}}
\newcommand{\tree}{%
\usebox{\treeboxone}
\setlength{\treeoffsetone}{\treeoffsettwo}%
\sbox{\treeboxone}{\usebox{\treeboxtwo}}%
\poptree}

\documentstyle[aclap,psfig]{article}

\newcommand{\derives}{\stackrel{\ast}{\Rightarrow}}

\title{Global Thresholding and Multiple-Pass
Parsing\thanks{\hspace{.3em}This material is based in part upon work
supported by the National Science Foundation under Grant
No. IRI-9350192 and a National Science Foundation Graduate Student
Fellowship.  I would also like to thank Michael Collins, Rebecca Hwa,
Lillian Lee, Wheeler Ruml, and Stuart Shieber for helpful discussions,
and comments on earlier drafts, and the anonymous reviewers for their
extensive comments.} }

\author{Joshua Goodman\\
        Harvard University\\
	40 Oxford St. \\
        Cambridge, MA 02138\\
        goodman@das.harvard.edu}

\begin{document}
\bibliographystyle{fullname}

\maketitle
\begin{abstract}
We present a variation on classic beam thresholding techniques that is
up to an order of magnitude faster than the traditional method, at the
same performance level.  We also present a new thresholding technique,
global thresholding, which, combined with the new beam thresholding,
gives an additional factor of two improvement, and a novel technique,
multiple pass parsing, that can be combined with the others to yield
yet another 50\% improvement.  We use a new search algorithm to
simultaneously optimize the thresholding parameters of the various
algorithms.
\end{abstract}

\section{Introduction}

In this paper, we examine thresholding techniques for statistical
parsers.  While there exist theoretically efficient ($O(n^3)$)
algorithms for parsing Probabilistic Context-Free Grammars (PCFGs) and
related formalisms, practical parsing algorithms usually make use of
pruning techniques, such as beam thresholding, for increased speed.

We introduce two novel thresholding techniques, global thresholding
and multiple-pass parsing, and one significant variation on
traditional beam thresholding.  We examine the value of these
techniques when used separately, and when combined.  In order to
examine the combined techniques, we also introduce an algorithm for
optimizing the settings of multiple thresholds.  When all three
thresholding methods are used together, they yield very
significant speedups over traditional beam thresholding, while
achieving the same level of performance.

We apply our techniques to CKY chart parsing, one of the most commonly
used parsing methods in natural language processing.  In a CKY chart
parser, a two-dimensional matrix of cells, the chart, is filled in.
Each cell in the chart corresponds to a span of the sentence, and each
cell of the chart contains the nonterminals that could generate that
span.  Cells covering shorter spans are filled in first, so we also
refer to this kind of parser as a bottom-up chart parser.

The parser fills in a cell in the chart by examining the nonterminals
in lower, shorter cells, and combining these nonterminals according to
the rules of the grammar.  The more nonterminals there are in the
shorter cells, the more combinations of nonterminals the parser must
consider.

In some grammars, such as PCFGs, probabilities are associated with the
grammar rules.  This introduces problems, since in many PCFGs, almost
any combination of nonterminals is possible, perhaps with some low
probability.  The large number of possibilities can greatly slow
parsing.  On the other hand, the probabilities also introduce new
opportunities.  For instance, if in a particular cell in the chart
there is some nonterminal that generates the span with high
probability, and another that generates that span with low
probability, then we can remove the less likely nonterminal from the
cell.  The less likely nonterminal will probably not be part of either
the correct parse or the tree returned by the parser, so removing it
will do little harm.  This technique is called {\em beam
thresholding}.

\begin{figure}
\begin{center}
\begin{tabular}{c}
\psfig{figure=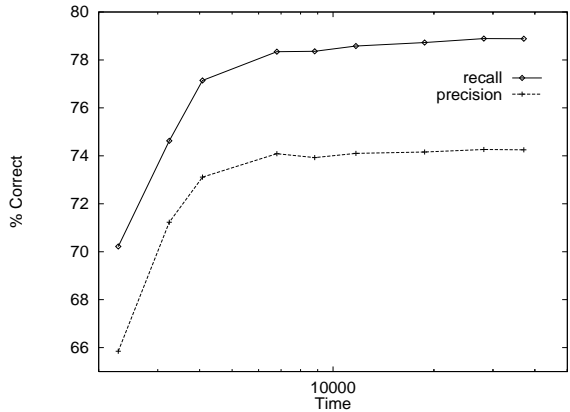,width=3.1in} 
\end{tabular}
\end{center}
\caption{Precision and Recall versus Time in Beam Thresholding}
\label{fig:tradeoff}
\end{figure}

If we use a loose beam threshold, removing only those nonterminals
that are much less probable than the best nonterminal in a cell, our
parser will run only slightly faster than with no thresholding, while
performance measures such as precision and recall will remain
virtually unchanged.  On the other hand, if we use a tight threshold,
removing nonterminals that are almost as probable as the
best nonterminal in a cell, then we can get a considerable speedup,
but at a considerable cost.  Figure \ref{fig:tradeoff} shows the
tradeoff between accuracy and time.

In this paper, we will consider three different kinds of thresholding.
The first of these is a variation on traditional beam search.  In
traditional beam search, only the probability of a nonterminal
generating the terminals of the cell's span is used.  We have found
that a minor variation, introduced in Section \ref{sec:beamthresh}, in
which we also consider the prior probability that each nonterminal is
part of the correct parse, can lead to nearly an order of magnitude
improvement.

The problem with beam search is that it only compares nonterminals to
other nonterminals in the same cell.  Consider the case in which a
particular cell contains only bad nonterminals, all of roughly equal
probability.  We can't threshold out these nodes, because even though
they are all bad, none is much worse than the best.  Thus, what we
want is a thresholding technique that uses some global information for
thresholding, rather than just using information in a single cell.
The second kind of thresholding we consider is a novel technique, {\em
global thresholding}, described in Section \ref{sec:globalthresh}.
Global thresholding makes use of the observation that for a
nonterminal to be part of the correct parse, it must be part of a
sequence of reasonably probable nonterminals covering the whole
sentence.

The last technique we consider, {\em multiple-pass parsing}, is
introduced in Section \ref{sec:multithresh}.  The basic idea is that
we can use information from parsing with one grammar to speed parsing
with another.  We run two passes, the first of which is fast and
simple, eliminating from consideration many unlikely potential
constituents.  The second pass is more complicated and slower, but
also more accurate.  Because we have already eliminated many nodes in
our first pass, the second pass can run much faster, and, despite the
fact that we have to run two passes, the added savings in the second
pass can easily outweigh the cost of the first one.

Experimental comparisons of these techniques show that they lead to
considerable speedups over traditional thresholding, when used
separately.  We also wished to combine the thresholding techniques;
this is relatively difficult, since searching for the optimal
thresholding parameters in a multi-dimensional space is potentially
very time consuming.  We designed a variant on a gradient descent
search algorithm to find the optimal parameters.  Using all three
thresholding methods together, and the parameter search algorithm, we
achieved our best results, running an estimated 30 times faster than
traditional beam search, at the same performance level.

\section{Beam Thresholding}

\label{sec:beamthresh}

The first, and simplest, technique we will examine is beam
thresholding.  While this technique is used as part of many search
algorithms, beam thresholding with PCFGs is most similar to beam
thresholding as used in speech recognition.  Beam thresholding is often
used in statistical parsers, such as that of \newcite{Collins:96a}.

Consider a nonterminal $X$ in a cell covering the span of terminals
$t_j...t_k$.  We will refer to this as {\em node} $N^X_{j,k}$, since
it corresponds to a potential node in the final parse tree.  Recall
that in beam thresholding, we compare nodes $N^X_{j,k}$ and
$N^Y_{j,k}$ covering the same span.  If one node is much more likely
than the other, then it is unlikely that the less probable node will
be part of the correct parse, and we can remove it from the chart,
saving time later.

There is a subtlety about what it means for a node $N^X_{j,k}$ to be
more likely than some other node.  According to folk wisdom, the best
way to measure the likelihood of a node $N^X_{j,k}$ is to use the
probability that the nonterminal $X$ generates the span $t_j...t_k$,
called the {\em inside probability}.  Formally, we write this as $P(X
\derives t_j...t_k)$, and denote it by $\beta(N^X_{j,k})$.  However,
this does not give information about the probability of the node in
the context of the full parse tree.  For instance, two nodes, one an
$\mathit{NP}$ and the other a $\mathit{FRAG}$ (fragment), may have
equal inside probabilities, but since there are far more
$\mathit{NP}$s than there are $\mathit{FRAG}$ clauses, the
$\mathit{NP}$ node is more likely overall.  Therefore, we must
consider more information than just the inside probability.

The {\em outside probability} of a node $N^X_{j,k}$ is the probability
of that node given the surrounding terminals of the sentence,
i.e. $P(S \derives t_1...t_{j-1} X t_{k+1} ... t_n)$, which we denote
by $\alpha(N^X_{j,k})$.  Ideally, we would multiply the inside
probability by the outside probability, and normalize.  This product
would give us the overall probability that the node is part of the
correct parse.  Unfortunately, there is no good way to quickly compute
the outside probability of a node during bottom-up chart parsing
(although it can be efficiently computed afterwards).  Thus, we
instead multiply the inside probability simply by the prior
probability of the nonterminal type, $P(X)$, which is an approximation
to the outside probability.  Our final thresholding measure is $P(X)
\times \beta(N^X_{j,k})$.  In Section \ref{sec:beamexp}, we will show
experiments comparing inside-probability beam thresholding to beam
thresholding using the inside probability times the prior.  Using the
prior can lead to a speedup of up to a factor of 10, at the same
performance level.

To the best of our knowledge, using the prior probability in beam
thresholding is new, although not particularly insightful on our part.
Collins (personal communication) independently observed the usefulness
of this modification, and \newcite{Caraballo:96a} used a related
technique in a best-first parser.  We think that the main reason this
technique was not used sooner is that beam thresholding for PCFGs is
derived from beam thresholding in speech recognition using Hidden
Markov Models (HMMs).  In an HMM, the forward probability of a given
state corresponds to the probability of reaching that state from the
start state.  The probability of eventually reaching the final state
from any state is always 1.  Thus, the forward probability is
all that is needed.  The same is true in some top down probabilistic
parsing algorithms, such as stochastic versions of Earley's algorithm
\cite{Stolcke:93a}.  However, in a bottom-up algorithm, we need the
extra factor that indicates the probability of getting from the start
symbol to the nonterminal in question, which we approximate by the
prior probability.  As we noted, this can be very different for
different nonterminals.

\section{Global Thresholding}

\label{sec:globalthresh}

As mentioned earlier, the problem with beam thresholding is that it
can only threshold out the worst nodes of a cell.  It cannot threshold
out an entire cell, even if there are no good nodes in it.  To remedy
this problem, we introduce a novel thresholding technique, global
thresholding.

\begin{figure}[t]
\psfig{figure=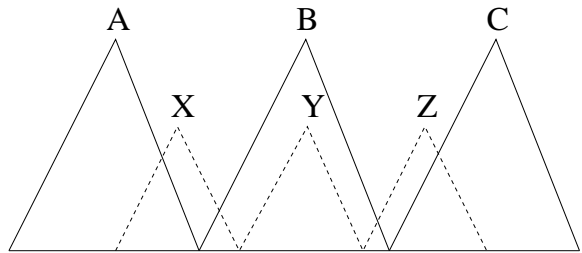,width=3in}
\caption{Global Thresholding Motivation} \label{fig:globalthresh}
\end{figure}

\newcommand{\alfor}{{\bf for }}
\newcommand{\alfloat}{{\bf float }}
\newcommand{\alif}{{\bf if }}
\newcommand{\mi}{\mathit}
\newcommand{\alforeach}{{\bf for each }}
\newcommand{\alelse}{{\bf else }}
\newcommand{\alto}{{\bf to }}
\newcommand{\alwhile}{{\bf while }}
\newcommand{\alnot}{{\bf not }}

\begin{figure}[t]
\setlength{\baselineskip}{0.7\baselineskip}

\begin{tabbing}
\verb|    |\=\verb|    |\=\verb|    |\=\verb|    |\=\verb|    |\=\verb|    |\=\verb|    |\=\verb|    |\=\verb|    |\=\verb|    |\= \kill
\alfloat $f[1..n\!+\!1] := \{1, 0, 0, ..., 0\}$; \\
\alfor $\mi{start}$ := 1 to $n$ \\
 \> \alforeach node $N$ beginning at $\mi{start}$ \\
 \>  \> $\mi{left} := f[\mi{start}];$ \\
 \>  \> $\mi{score} := \mi{left} \times N_{\mi{inside}} \times N_{\mi{prior}}$; \\
 \>  \> \alif $\mi{score} > f[\mi{start} + N_{\mi{length}}]$ \\
 \>  \>  \> $f[\mi{start} + N_{\mi{length}}] := \mi{score}$; \\
\\
\alfloat $b[1..n\!+\!1] := \{0, ..., 0, 0, 1\}$; \\
\alfor $\mi{start}$ := $n$ downto 1 \\
 \> \alforeach node $N$ beginning at $\mi{start}$ \\
 \>  \> $\mi{right} := b[\mi{start} + N_{\mi{length}}]$; \\
 \>  \> $\mi{score} := \mi{right} \times N_{\mi{inside}} \times N_{\mi{prior}}$; \\
 \>  \> \alif $\mi{score} > b[\mi{start}]$ \\
 \>  \>  \> $b[\mi{start}] := \mi{score}$; \\
\\
$\mi{bestProb} := f[n\!+\!1]$; \\
\\
\alforeach node $N$ \\
 \> $\mi{left} := f[N_{\mi{start}}]$; \\
 \> $\mi{right} := b[N_{\mi{start}} + N_{\mi{length}}]$; \\
 \> $\mi{total} := \mi{left} \times N_{\mi{inside}} \times N_{\mi{prior}} \times \mi{right}$; \\
 \> \alif $\mi{total} > \mi{bestProb} \times T_G$ \\
 \>  \>  $N_{\mi{active}} := $ {\bf TRUE}; \\
 \> \alelse \\
 \>  \>  $N_{\mi{active}} := $ {\bf FALSE}; \\
 
\end{tabbing}
\caption{Global Thresholding Algorithm} \label{fig:globalg}
\end{figure}

\newcommand{\etal}{{et al.\ }}

The key insight of global thresholding is due to \newcite{Rayner:96a}.
Rayner \etal noticed that a particular node cannot be part of
the correct parse if there are no nodes in adjacent cells.  In
fact, it must be part of a sequence of nodes stretching from the start
of the string to the end.  In a probabilistic framework where almost
every node will have some (possibly very small) probability, we can
rephrase this requirement as being that the node must be part of a
reasonably probable sequence.  

Figure \ref{fig:globalthresh} shows an example of this insight.
Nodes A, B, and C will not be thresholded out, because each is part of
a sequence from the beginning to the end of the chart.  On the other
hand, nodes X, Y, and Z will be thresholded out, because none is part
of such a sequence.

Rayner \etal used this insight for a hierarchical, non-recursive
grammar, and only used their technique to prune after the first level
of the grammar.  They computed a score for each sequence as the
minimum of the scores of each node in the sequence, and computed a
score for each node in the sequence as the minimum of three scores:
one based on statistics about nodes to the left, one based on nodes to
the right, and one based on unigram statistics.

We wanted to extend the work of Rayner \etal to general PCFGs,
including those that were recursive.  Our approach therefore differs
from theirs in many ways.  Rayner \etal ignore the inside
probabilities of nodes; while this may work after processing only the
first level of a grammar, when the inside probabilities will be
relatively homogeneous, it could cause problems after other levels,
when the inside probability of a node will give important information
about its usefulness.  On the other hand, because long nodes will tend
to have low inside probabilities, taking the minimum of all scores
strongly favors sequences of short nodes.  Furthermore, their
algorithm requires time $O(n^3)$ to run just once.  This is acceptable
if the algorithm is run only after the first level, but running it
more often would lead to an overall run time of $O(n^4)$.  Finally, we
hoped to find an algorithm that was somewhat less heuristic in nature.

Our global thresholding technique thresholds out node $N$ if the ratio
between the most probable sequence of nodes including node $N$ and the
overall most probable sequence of nodes is less than some threshold,
$T_G$.  Formally, denoting sequences of nodes by $L$, we threshold
node $N$ if
$$
T_G \max_{L} P(L) > \max_{L | N \in L} P(L)
$$

Now, the hard part is determining $P(L)$, the probability of a node
sequence.  Unfortunately, there is no way to do this efficiently as
part of the intermediate computation of a bottom-up chart
parser.\footnote{Some other parsing techniques, such as stochastic
versions of Earley parsers \cite{Stolcke:93a}, efficiently compute
related probabilities, but we won't explore these parsers here.  We
confess that our real interest is in more complicated grammars, such
as those that use head words.  Grammars such as these can best be
parsed bottom up.}  We will
approximate $P(L)$ as follows:
$$
P(L) = \prod_{i} P(L_i|L_1...L_{i-1}) \approx \prod_{i} P(L_i)
$$
That is, we assume independence between the elements of a sequence.
The probability of node $L_i=N^X_{j,k}$ is just its prior probability
times its inside probability, as before.

The most important difference between global thresholding and beam
thresholding is that global thresholding is global: any node in the
chart can help prune out any other node.  In stark contrast, beam
thresholding only compares nodes to other nodes covering the same
span.  Beam thresholding typically allows tighter thresholds since
there are fewer approximations, but does not benefit from global
information.

\subsection{Global Thresholding Algorithm}
Global thresholding is performed in a bottom-up chart parser
immediately after each length is completed.  It thus runs $n$ times
during the course of parsing a sentence of length $n$.

We use the simple dynamic programming algorithm in Figure
\ref{fig:globalg}.  There are $O(n^2)$ nodes in the chart, and each
node is examined exactly three times, so the run time of this
algorithm is $O(n^2)$.  The first section of the algorithm works
forwards, computing, for each $i$, $f[i]$, which contains the score of
the best sequence covering terminals $t_1...t_{i-1}$.  Thus
$f[n\!+\!1]$ contains the score of the best sequence covering the
whole sentence, $\max_{L} P(L)$.  The algorithm works analogously to
the Viterbi algorithm for HMMs.  The second section is analogous, but
works backwards, computing $b[i]$, which contains the score of the
best sequence covering terminals $t_i...t_n$. 

Once we have computed the preceding arrays, computing $\max_{L | N \in
L} P(L)$ is straightforward.  We simply want the score of the best
sequence covering the nodes to the left of $N$, $f[N_{\mi{start}}]$,
times the score of the node itself, times the score of the best
sequence of nodes from $N_{\mi{start}} + N_{\mi{length}}$ to the end,
which is just $b[N_{\mi{start}} + N_{\mi{length}}]$.  Using this
expression, we can threshold each node quickly.

Since this algorithm is run $n$ times during the course of parsing,
and requires time $O(n^2)$ each time it runs, the algorithm requires
time $O(n^3)$ overall.  Experiments will show that the time it saves
easily outweighs the time it uses.

\section{Multiple-Pass Parsing}

\label{sec:multithresh}

In this section, we discuss a novel thresholding technique,
multiple-pass parsing.  We show that multiple-pass parsing techniques
can yield large speedups.  Multiple-pass parsing is a variation on a
new technique in speech recognition, multiple-pass speech recognition
\cite{Zavaliagkos:94a}, which we introduce first.

\subsection{Multiple-Pass Speech Recognition}

In an idealized multiple-pass speech recognizer, we first run a simple
pass, computing the forward and backward probabilities.  This first
pass runs relatively quickly.  We can use information from this
simple, fast first pass to eliminate most states, and then run a more
complicated, slower second pass that does not examine states that
were deemed unlikely by the first pass.  The extra time of running two
passes is more than made up for by the time saved in the second pass.

The mathematics of multiple-pass recognition is fairly simple.  In the
first simple pass, we record the forward probabilities,
$\alpha(S^t_i)$, and backward probabilities, $\beta(S^t_i)$, of each
state $i$ at each time $t$.  Now, $\frac{\alpha(S^t_i) \times
\beta(S^t_i)}{\alpha(S^T_{\mathit{final}})}$ gives the overall
probability of being in state $i$ at time $t$ given the acoustics.
Our second pass will use an HMM whose states are analogous to the
first pass HMM's states.  If a first pass state at some time is
unlikely, then the analogous second pass state is probably also
unlikely, so we can threshold it out.

There are a few complications to multiple-pass recognition.  First,
storing all the forward and backward probabilities can be expensive.
Second, the second pass is more complicated than the first, typically
meaning that it has more states.  So the mapping between states in the
first pass and states in the second pass may be non-trivial.  To
solve both these problems, only states at word transitions are saved.
That is, from pass to pass, only information about where words are
likely to start and end is used for thresholding.

\subsection{Multiple-Pass Parsing}

We can use an analogous algorithm for multiple-pass parsing.  In
particular, we can use two grammars, one fast and simple and the other
slower, more complicated, and more accurate.  Rather than using the
forward and backward probabilities of speech recognition, we use the
analogous inside and outside probabilities, $\beta(N^X_{j,k})$ and
$\alpha(N^X_{j,k})$ respectively.  Remember that
$\frac{\alpha(N^X_{j,k}) \beta(N^X_{j,k})}{\beta(N^S_{1,n})}$ is the
probability that $N^X_{j,k}$ is in the correct parse (given, as always,
the model and the string).  Thus, we run our first pass, computing
this expression for each node.  We can then eliminate from
consideration in our later passes all nodes for which the probability
of being in the correct parse was too small in the first pass.

Of course, for our second pass to be more accurate, it will probably
be more complicated, typically containing an increased number of
nonterminals and productions.  Thus, we create a mapping function from
each first pass nonterminal to a set of second pass nonterminals, and
threshold out those second pass nonterminals that map from low-scoring
first pass nonterminals.  We call this mapping function the {\em
descendants function}.\footnote{In this paper, we will assume that
each second pass nonterminal can descend from at most one first pass
nonterminal in each cell.  The grammars used here have this property.
If this assumption is violated, multiple-pass parsing is still
possible, but some of the algorithms need to be changed.}

There are many possible examples of first and second pass
combinations.  For instance, the first pass could use regular
nonterminals, such as $\mathit{NP}$ and $\mathit{VP}$ and the second
pass could use nonterminals augmented with head-word information.  The
descendants function then appends the possible head words to the first
pass nonterminals to get the second pass ones.

\begin{figure}
\begin{tabbing}
\verb|  |\=\verb|  |\=\verb|  |\=\verb|  |\=\verb|  |\=\verb|  |\=\verb|  |\=\verb|  |\=\verb|  |\=\verb|  |\= \kill
\alfor $\mi{length} := 2$ \alto $n$ \\
 \> \alfor $\mi{start} := 1$ \alto $n-\mi{length}+1$ \\
 \> \> \alfor $\mi{leftLength} := 1$ \alto $\mi{length}-1$ \\
 \> \>  \> $\mi{LeftPrev} := \mi{PrevChart}[\mi{leftLength}][\mi{start}]$; \\
 \> \>  \> \alforeach $\mi{LeftNodePrev} \in \mi{LeftPrev}$ \\
 \> \>  \>  \> \alforeach production instance $\mi{Prod}$ from \\
 \> \>  \>  \>  \>  \>  \> $\mi{LeftNodePrev}$ of size $\mi{length}$ \\
 \> \>  \>  \>  \> \alforeach descendant $L$ of $\mi{Prod}_{\mi{Left}}$ \\
 \> \>  \>  \>  \>  \> \alforeach descendant $R$ of $\mi{Prod}_{\mi{Right}}$ \\
 \> \>  \>  \>  \>  \>  \> \alforeach descendant $P$ of $\mi{Prod}_{\mi{Parent}}$ \\
 \> \>  \>  \>  \>  \>  \>  \>  \>  \> such that $P \rightarrow L\; R$ \\
 \> \>  \>  \>  \>  \>  \>  \> add $P$ to $\mi{Chart}[\mathit{length}][\mi{start}]$; \\
\end{tabbing}
\caption{Second Pass Parsing Algorithm} \label{fig:secondpass}
\end{figure}


Even though the correspondence between forward/backward and
inside/outside probabilities is very close, there are important
differences between speech-recognition HMMs and natural-language
processing PCFGs.  In particular, we have found that it is more
important to threshold productions than nonterminals.  That is, rather
than just noticing that a particular nonterminal $\mathit{VP}$
spanning the words ``killed the rabbit'' is very likely, we also note
that the production $ \mathit{VP} \rightarrow V\;\mathit{NP}$ (and the
relevant spans) is likely.

Both the first and second pass parsing algorithms are simple
variations on CKY parsing.  In the first pass, we now keep track of
each {\em production instance} associated with a node, i.e. $N^X_{i,j}
\rightarrow N^Y_{i,k}\;N^Z_{k+1,j}$, computing the inside and outside
probabilities of each.  The second pass requires more changes.  Let us
denote the descendants of nonterminal $X$ by $X_1...X_x$.  In the
second pass, for each production of the form $N^X_{i,j} \rightarrow
N^Y_{i,k}\;N^Z_{k+1,j}$ in the first pass that wasn't thresholded out
by multi-pass thresholding, beam thresholding, etc., we consider every
descendant production instance, that is, all those of the form
$N^{X_p}_{i,j} \rightarrow N^{Y_q}_{i,k}\;N^{Z_r}_{k+1,j}$, for
appropriate values of $p, q, r$.  This algorithm is given in Figure
\ref{fig:secondpass}, which uses a current pass matrix $\mi{Chart}$ to
keep track of nonterminals in the current pass, and a previous pass
matrix, $\mi{PrevChart}$ to keep track of nonterminals in the previous
pass.  We use one additional optimization, keeping track of the
descendants of each nonterminal in each cell in $\mi{PrevChart}$ which
are in the corresponding cell of $\mi{Chart}$.

We tried multiple-pass thresholding in two different ways.  In the
first technique we tried, production-instance thresholding, we remove
from consideration in the second pass the descendants of all
production instances whose combined inside-outside probability falls
below a threshold.  In the second technique, node thresholding, we
remove from consideration the descendants of all nodes whose
inside-outside probability falls below a threshold.  In our pilot
experiments, we found that in some cases one technique works slightly
better, and in some cases the other does.  We therefore ran our
experiments using both thresholds together.

One nice feature of multiple-pass parsing is that under special
circumstances, it is an {\em admissible} search technique, meaning
that we are guaranteed to find the best solution with it.  In
particular, if we parse using no thresholding, and our grammars have
the property that for every non-zero probability parse in the second
pass, there is an analogous non-zero probability parse in the first
pass, then multiple-pass search is admissible.  Under these
circumstances, no non-zero probability parse will be thresholded out,
but many zero probability parses may be removed from consideration.
While we will almost always wish to parse using thresholds, it is nice
to know that multiple-pass parsing can be seen as an approximation to
an admissible technique, where the degree of approximation is
controlled by the thresholding parameter.

\section{Multiple Parameter Optimization}

\label{sec:descent}

The use of any one of these techniques does not exclude the use of the
others.  There is no reason that we cannot use beam thresholding,
global thresholding, and multiple-pass parsing all at the same time.
In general, it wouldn't make sense to use a technique such as
multiple-pass parsing without other thresholding techniques; our
first pass would be overwhelmingly slow without some sort of
thresholding.

There are, however, some practical considerations.  To optimize a
single threshold, we could simply sweep our parameters over a one
dimensional range, and pick the best speed versus performance
tradeoff.  In combining multiple techniques, we need to find optimal
combinations of thresholding parameters.  Rather than having to
examine 10 values in a single dimensional space, we might have to
examine 100 combinations in a two dimensional space.  Later, we show
experiments with up to six thresholds.  Since we don't have time to
parse with one million parameter combinations, we need a better search
algorithm.

\begin{figure}
\setlength{\baselineskip}{0.7\baselineskip}
\begin{tabbing}
\verb|  |\=\verb|  |\=\verb|  |\=\verb|  |\=\verb|  |\=\verb|  |\=\verb|  |\=\verb|  |\=\verb|  |\=\verb|  |\= \kill
\alwhile \alnot $\mi{Thresholds} \in \mi{ThresholdsSet}$ \\
 \> add $\mi{Thresholds}$ to $\mi{ThresholdsSet};$ \\
 \> $(\mi{BaseE_T}, \mi{BaseTime}) :=$ ParseAll$(\mi{Thresholds})$; \\
 \> \alforeach $\mi{Threshold} \in \mi{Thresholds}$ \\
 \>  \> if $\mi{BaseE_T} > \mi{TargetE_T}$ \\
 \>  \>  \> tighten $\mi{Threshold}$; \\
 \>  \>  \> $(\mi{NewE_T}, \mi{NewTime}) :=$ ParseAll$(\mi{Thresholds})$; \\
 \>  \>  \> $\mi{Ratio} := (\mi{BaseTime}-\mi{NewTime}) \; / $ \\
 \>  \>  \>  \>  \>  \>  $(\mi{BaseE_T}-\mi{NewE_T})$; \\
 \>  \> \alelse \\
 \>  \>  \> loosen $\mi{Threshold}$; \\
 \>  \>  \> $(\mi{NewE_T}, \mi{NewTime}) :=$ ParseAll$(\mi{Thresholds})$; \\
 \>  \>  \> $\mi{Ratio} := (\mi{BaseE_T}-\mi{NewE_T})\;/$ \\
 \>  \>  \>  \>  \>  \> $(\mi{BaseTime}-\mi{NewTime})$; \\
 \> change $\mi{Threshold}$ with best $\mi{Ratio}$; \\
\end{tabbing}

\caption{Gradient Descent Multiple Threshold Search} \label{fig:multisearch}
\end{figure}

\begin{figure}
\begin{center}
\begin{tabular}{cc}
\psfig{figure=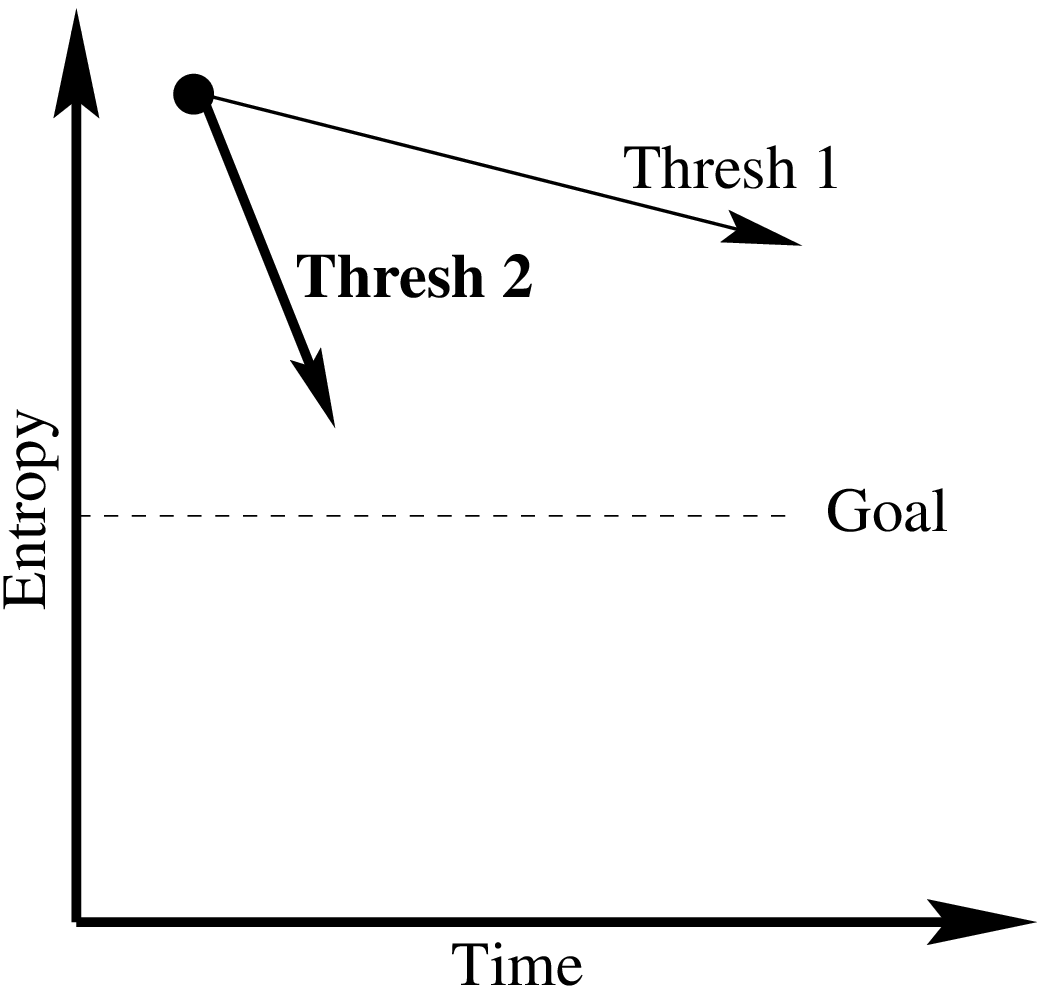,width=1.4in}  &
\psfig{figure=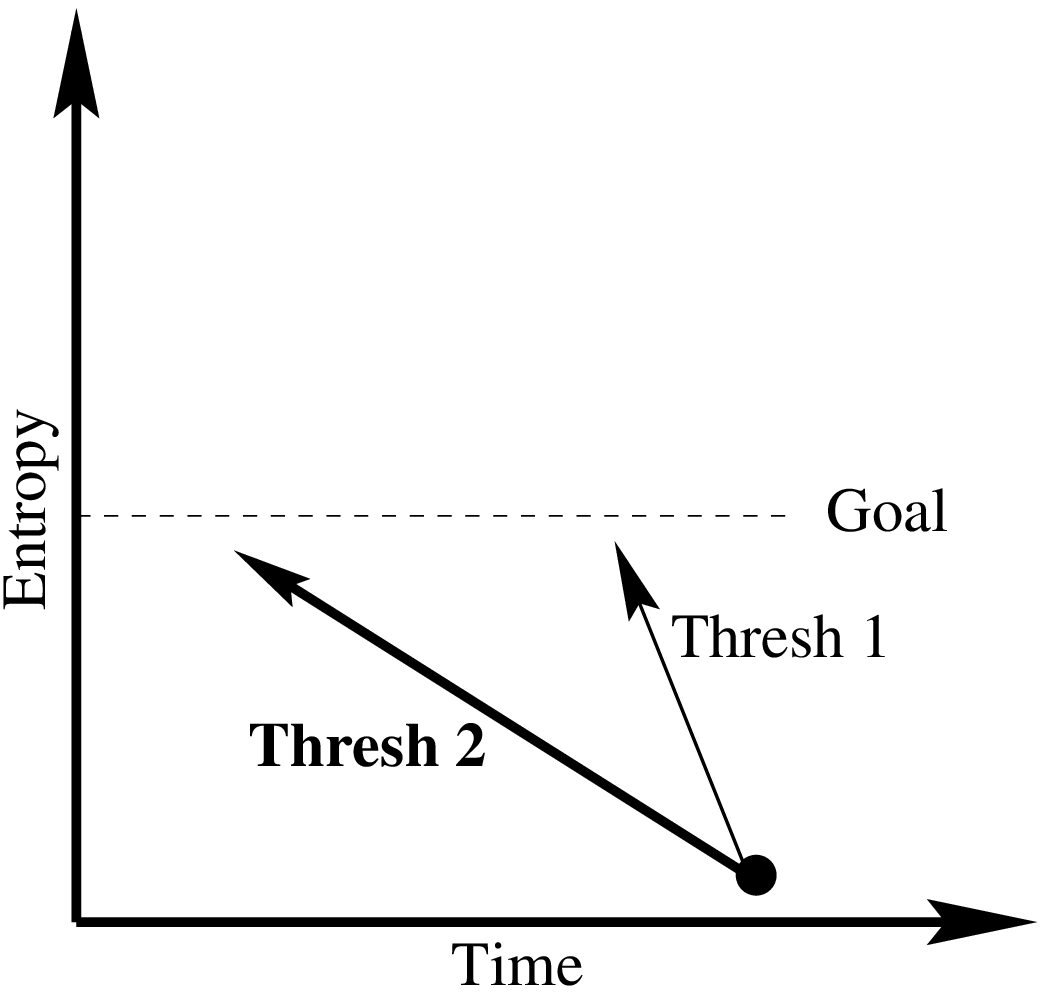,width=1.4in} \\
Optimizing for & Optimizing for \\
{\bf Lower Entropy}: &  {\bf Faster Speed}: \\
{\bf Steeper} is Better &  {\bf Flatter} is Better \\
\end{tabular}
\caption{Optimizing for Lower Entropy versus Optimizing for Faster
Speed} \label{fig:optimpic}
\end{center}
\end{figure}

Ideally, we would like to be able to pick a performance level (in
terms of either entropy or precision and recall) and find the best set
of thresholds for achieving that performance level as quickly as
possible.  If this is our goal, then a normal gradient descent
technique won't work, since we can't use such a technique to optimize
one function of a set of variables (time as a function of thresholds)
while holding another one constant (performance).\footnote{We could
use gradient descent to minimize a weighted sum of time and
performance, but we wouldn't know at the beginning what performance we
would have at the end.  If our goal is to have the best performance we
can while running in real time, or to achieve a minimum acceptable
performance level with as little time as necessary, then a simple
gradient descent function wouldn't work as well as our algorithm.

Also, for this algorithm (although not for most experiments), our
measurement of time was the total number of productions searched,
rather than cpu time; we wanted the greater accuracy of measuring
productions.}

We wanted a metric of performance which would be sensitive to changes
in threshold values.  In particular, our ideal metric would be
strictly increasing as our thresholds loosened, so that every
loosening of threshold values would produce a measurable increase in
performance.  The closer we get to this ideal, the fewer sentences we
need to test during parameter optimization.

We tried an experiment in which we ran beam thresholding with a tight
threshold, and then a loose threshold, on all sentences of section 0
of length $\le 40$.  For this experiment only, we discarded those
sentences which could not be parsed with the specified setting of the
threshold, rather than retrying with looser thresholds.  We then
computed for each of six metrics how often the metric decreased,
stayed the same, or increased for each sentence between the two runs.
Ideally, as we loosened the threshold, every sentence should improve
on every metric, but in practice, that wasn't the case.  As can be
seen, the inside score was by far the most nearly strictly increasing
metric.  Therefore, we should use the inside probability as our metric
of performance; however inside probabilities can become very close to
zero, so instead we measure entropy, the negative logarithm of the
inside probability.

\begin{tabular}{r|lll}
Metric & decrease & same & increase \\ \hline
              Inside &    7 &   65 & 1625 \\
             Viterbi &    6 & 1302 &  389 \\
       Cross Bracket &  132 & 1332 &  233 \\
  Zero Cross Bracket &   18 & 1616 &   63 \\
           Precision &  132 & 1280 &  285 \\
              Recall &  126 & 1331 &  240 \\
\end{tabular}


We implemented a variation on a steepest descent search technique.  We
denote the entropy of the sentence after thresholding by $E_T$.  Our
search engine is given a target performance level $E_T$ to search for,
and then tries to find the best combination of parameters that works
at approximately this level of performance.  At each point, it finds
the threshold to change that gives the most ``bang for the buck.''  It
then changes this parameter in the correct direction to move towards
$E_T$ (and possibly overshoot it).  A simplified version of the
algorithm is given in Figure \ref{fig:multisearch}.

Figure \ref{fig:optimpic} shows graphically how the algorithm works.
There are two cases.  In the first case, if we are currently above the
goal entropy, then we loosen our thresholds, leading to slower speed
and lower entropy.  We then wish to get as much entropy reduction as
possible per time increase; that is, we want the steepest slope possible.
On the other hand, if we are trying to increase our entropy, we want
as much time decrease as possible per entropy increase; that is, we want
the flattest slope possible.  Because of this difference, we need to
compute different ratios depending on which side of the goal we are on.

There are several subtleties when thresholds are set very tightly.
When we fail to parse a sentence because the thresholds are too tight,
we retry the parse with lower thresholds.  This can lead to conditions
that are the opposite of what we expect; for instance, loosening
thresholds may lead to faster parsing, because we don't need to parse
the sentence, fail, and then retry with looser thresholds.  The full
algorithm contains additional checks that our thresholding change had
the effect we expected (either increased time for decreased entropy or
vice versa).  If we get either a change in the wrong direction, or a
change that makes everything worse, then we retry with the inverse
change, hoping that that will have the intended effect.  If we get a
change that makes both time and entropy better, then we make that
change regardless of the ratio.

Also, we need to do checks that the denominator when computing $Ratio$
isn't too small.  If it is very small, then our estimate may be
unreliable, and we don't consider changing this parameter.  Finally,
the actual algorithm we used also contained a simple ``annealing
schedule'', in which we slowly decreased the factor by which we changed
thresholds.  That is, we actually run the algorithm multiple times to
termination, first changing thresholds by a factor of 16.  After a
loop is reached at this factor, we lower the factor to 4,
then 2, then 1.414, then 1.15.

Note that this algorithm is fairly domain independent.  It can be used
for almost any statistical parsing formalism that uses thresholds, or
even for speech recognition.

\section{Comparison to Previous Work}

Beam thresholding is a common approach.  While we don't know of other
systems that have used exactly our techniques, our techniques are
certainly similar to those of others.  For instance,
\newcite{Collins:96a} uses a form of beam thresholding that differs
from ours only in that it doesn't use the prior probability of
nonterminals as a factor, and \newcite{Caraballo:96a} use a version
with the prior, but with other factors as well.

Much of the previous related work on thresholding is in the similar
area of priority functions for agenda-based parsers.  These parsers
try to do ``best first'' parsing, with some function akin to a
thresholding function determining what is best.  The best comparison
of these functions is due to Caraballo and Charniak
\shortcite{Caraballo:96a,Caraballo:97a}, who tried various
prioritization methods.  Several of their techniques are similar to
our beam thresholding technique, and one of their techniques, not yet
published \cite{Caraballo:97a}, would probably work better.

The only technique that Caraballo and Charniak
\shortcite{Caraballo:96a} give that took into account the scores of
other nodes in the priority function, the ``prefix model,'' required
$O(n^5)$ time to compute, compared to our $O(n^3)$ system.  On the
other hand, all nodes in the agenda parser were compared to all other
nodes, so in some sense all the priority functions were global.

Note that agenda-based PCFG parsers in general require more than
$O(n^3)$ run time, because, when better derivations are discovered,
they may be forced to propagate improvements to productions that they
have previously considered.  For instance, if an agenda-based system
first computes the probability for a production $S \rightarrow
\mathit{NP}\;\mathit{VP}$, and then later computes some better
probability for the $\mathit{NP}$, it must update the probability for
the $S$ as well.  This could propagate through much of the chart.  To
remedy this, Caraballo \etal only propagated probabilities that
caused a large enough change \cite{Caraballo:97a}.  Also, the question
of when an agenda-based system should stop is a little discussed
issue, and difficult since there is no obvious stopping criterion.
Because of these issues, we chose not to implement an agenda-based
system for comparison.

As mentioned earlier, \newcite{Rayner:96a} describe a system that is
the inspiration for global thresholding.  Because of the limitation of
their system to non-recursive grammars, and the other differences
discussed in Section \ref{sec:globalthresh}, global thresholding represents
a significant improvement.

\newcite{Collins:96a} uses two thresholding techniques.  The first of
these is essentially beam thresholding without a prior.  In the second
technique, there is a constant probability threshold.  Any nodes with
a probability below this threshold are pruned.  If the parse fails,
parsing is restarted with the constant lowered.  We attempted to
duplicate this technique, but achieved only negligible performance
improvements.  Collins (personal communication) reports a 38\% speedup
when this technique is combined with loose beam thresholding, compared to
loose beam thresholding alone.  Perhaps our lack of success is due to
differences between our grammars, which are fairly different
formalisms.  When Collins began using a formalism somewhat closer to
ours, he needed to change his beam thresholding to take into account
the prior, so this is not unlikely.  Hwa (personal
communication) using a model similar to PCFGs, Stochastic Lexicalized
Tree Insertion Grammars, also was not able to obtain a speedup using
this technique.

There is previous work in the speech recognition community on
automatically optimizing some parameters \cite{Schwartz:92a}.
However, this previous work differed significantly from ours both in
the techniques used, and in the parameters optimized.  In particular,
previous work focused on optimizing weights for various components,
such as the language model component.  In contrast, we optimize
thresholding parameters.  Previous techniques could not be used or
easily adapted to thresholding parameters.

\section{Experiments}

\subsection{The Parser and Data}

\begin{figure}[t]
\setlength{\baselineskip}{0.7\baselineskip}
\begin{tabbing}
\verb|    |\=\verb|    |\=\verb|    |\=\verb|    |\=\verb|    |\=\verb|    |\=\verb|    |\=\verb|    |\=\verb|    |\=\verb|    |\= \kill

\alforeach rule $P \rightarrow L\;R$ \\
 \> \alif nonterminal $L$ in left cell \\
 \>  \> \alif nonterminal $R$ in right cell \\
 \>  \>  \> add $P$ to parent cell; \\
 \\
Algorithm One \\
\\
\alforeach nonterminal $L$ in left cell \\
 \> \alforeach nonterminal $R$ in right cell \\
 \>  \> \alforeach rule $P \rightarrow L\; R$ \\
 \>  \>  \> add $P$ to parent cell; \\
 \\
Algorithm Two \\
\end{tabbing}

\caption{Two Possible CKY inner loops} \label{fig:twoinner}
\end{figure}

\begin{figure}[htb]
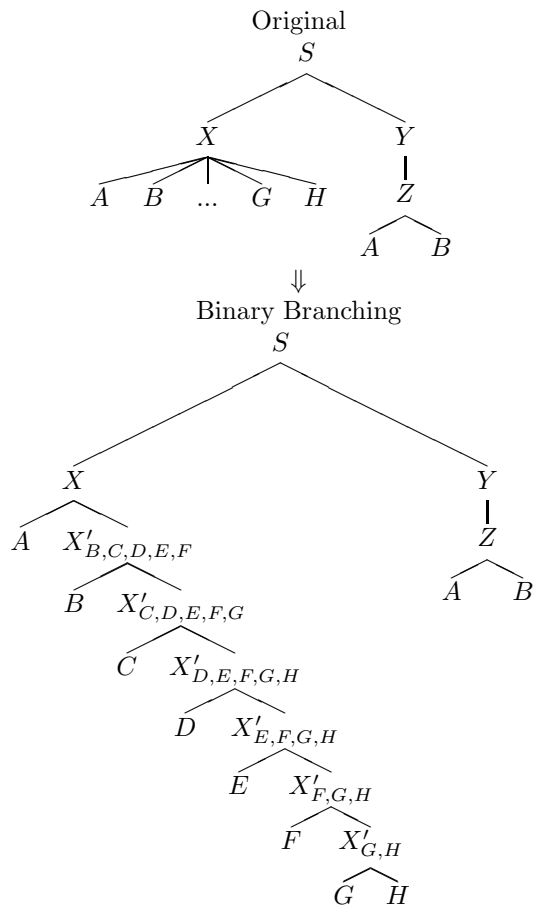

\begin{center}
\begin{tabular}{c}
Original \\
\leaf{$A$}
\leaf{$B$}
\leaf{$...$}
\leaf{$G$}
\leaf{$H$}
\branch{5}{$X$}
\leaf{$A$}\faketreewidth{AA}
\leaf{$B$}\faketreewidth{AA}
\branch{2}{$Z$}
\branch{1}{$Y$}
\branch{2}{$S$}
\hspace{-.8in}\tree \\
$\Downarrow$ \\
Binary Branching \\
\leaf{$A$}
\leaf{$B$}
\leaf{$C$}
\leaf{$D$}
\leaf{$E$}
\leaf{$F$}
\leaf{$G$}
\leaf{$H$}
\branch{2}{$X^{\prime}_{G,H}$}
\branch{2}{$X^{\prime}_{F,G,H}$}
\branch{2}{$X^{\prime}_{E,F,G,H}$}
\branch{2}{$X^{\prime}_{D,E,F,G,H}$}
\branch{2}{$X^{\prime}_{C,D,E,F,G}$}
\branch{2}{$X^{\prime}_{B,C,D,E,F}$}
\branch{2}{$X$}
\leaf{$A$}\faketreewidth{AA}
\leaf{$B$}\faketreewidth{AA}
\branch{2}{$Z$}
\branch{1}{$Y$}
\branch{2}{$S$}
\hspace {-1.2in}\tree
\end{tabular}
\end{center}
\caption{Converting to Binary Branching}
\label{fig:binary}
\end{figure}

The inner loop of the CKY algorithm, which determines for every pair
of cells what nodes must be added to the parent, can be written in
several different ways.  Which way this is done interacts with
thresholding techniques.  There are two possibilities, as shown in
Figure \ref{fig:twoinner}.  We used the second technique, since the
first technique gets no speedup from most thresholding systems.

All experiments were trained on sections 2-18 of the Penn Treebank,
version II.  A few were tested, where noted, on the first 200
sentences of section 00 of length at most 40 words.  In one
experiment, we used the first 15 of length at most 40, and in the
remainder of our experiments, we used those sentences in the first
1001 of length at most 40.  Our parameter optimization algorithm
always used the first 31 sentences of length at most 40 words from
section 19.  We ran some experiments on more sentences, but there were
three sentences in this larger test set that could not be parsed with
beam thresholding, even with loose settings of the threshold; we
therefore chose to report the smaller test set, since it is difficult
to compare techniques which did not parse exactly the same sentences.

\subsection{The Grammar}

We needed several grammars for our experiments so that we could test
the multiple-pass parsing algorithm.  The grammar rules, and their
associated probabilities, were determined by reading them off of the
training section of the treebank, in a manner very similar to that
used by \newcite{Charniak:96a}.  The main grammar we chose was
essentially of the following form:

\begin{center}
\renewcommand{\arraystretch}{0.5}
\begin{tabular}{rcl}
$X$                  & $\Rightarrow$ & $A \; X^{\prime}_{B,C,D,E,F} $ \\
$X^{\prime}_{A,B,C,D,E}$      & $\Rightarrow$ & $A \; X^{\prime}_{B,C,D,E,F} $ \\
$X$                  & $\Rightarrow$ & $A$ \\
$X$                  & $\Rightarrow$ & $A \; B$ \\
\end{tabular}
\end{center}

That is, our grammar was binary branching except that we also allowed
unary branching productions.  There were never more than five
subscripted symbols for any nonterminal, although there could be fewer
than five if there were fewer than five symbols remaining on the right
hand side.  Thus, our grammar was a kind of 6-gram model on symbols in
the grammar.\footnote{We have skipped over details regarding our
handling of unary branching nodes.  Unary branching nodes are in
general difficult to deal with \cite{Stolcke:93a}.  The actual
grammars we used contained additional symbols in such a way that there
could not be more than one unary branch in a row.  This greatly
simplified computations, especially of the inside and outside
probabilities.  We also doubled the number of cells in our parser,
having both unary and binary cells for each length/start pair.}
Figure \ref{fig:binary} shows an example of how we converted trees to
binary branching with our grammar.  We refer to this grammar as the
{\em 6-gram grammar}.  The terminals of the grammar were the
part-of-speech symbols in the treebank.  Any experiments that don't
mention which grammar we used were run with the 6-gram grammar.

\begin{figure}[htb]
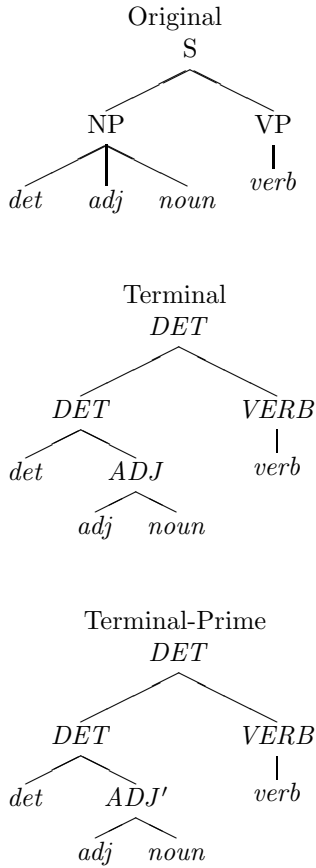

\begin{center}
\begin{tabular}{c}
Original \\
\leaf{$\mathit{det}$}
\leaf{$\mathit{adj}$}
\leaf{$\mathit{noun}$}
\branch{3}{NP}
\leaf{$\mathit{verb}$}
\branch{1}{VP}
\branch{2}{S}
\hspace{-.6in} \tree  \\ \\ \\
Terminal \\
\leaf{$\mathit{det}$}
\leaf{$\mathit{adj}$}
\leaf{$\mathit{noun}$}
\branch{2}{$\mathit{ADJ}$}
\branch{2}{$\mathit{DET}$}
\leaf{$\mathit{verb}$}
\branch{1}{$\mathit{VERB}$}
\branch{2}{$\mathit{DET}$}
\hspace{-.55in} \tree \\ \\ \\
Terminal-Prime \\
\leaf{$\mathit{det}$}
\leaf{$\mathit{adj}$}
\leaf{$\mathit{noun}$}
\branch{2}{$\mathit{ADJ'}$}
\branch{2}{$\mathit{DET}$}
\leaf{$\mathit{verb}$}
\branch{1}{$\mathit{VERB}$}
\branch{2}{$\mathit{DET}$}
\hspace{-.55in} \tree 
\\
\end{tabular}
\end{center}
\caption{Converting to Terminal and Terminal-Prime Grammars 
\label{fig:maketerm}}
\end{figure}

For a simple grammar, we wanted something that would be very fast.
The fastest grammar we can think of we call the {\em terminal}
grammar, because it has one nonterminal for each terminal symbol in
the alphabet.  The nonterminal symbol indicates the first terminal in
its span.  The parses are binary branching in the same way that the
6-gram grammar parses are.  Figure \ref{fig:maketerm} shows how to
convert a parse tree to the terminal grammar.  Since there is only one
nonterminal possible for each cell of the chart, parsing is quick for
this grammar.  For technical and practical reasons, we actually wanted
a marginally more complicated grammar, which included the ``prime''
symbol of the 6-gram grammar, indicating that a cell is part of the
same constituent as its parent.  Therefore, we doubled the size of the
grammar so that there would be both primed and non-primed versions of
each terminal; we call this the {\em terminal-prime} grammar, and also
show how to convert to it in Figure \ref{fig:maketerm}.  This is the
grammar we actually used as the first pass in our multiple-pass
parsing experiments.

\subsection{What we measured}

\label{sec:expmeasure}

The goal of a good thresholding algorithm is to trade off correctness
for increased speed.  We must thus measure both correctness and speed,
and there are some subtleties to measuring each.

First, the traditional way of measuring correctness is with metrics
such as precision and recall.  Unfortunately, there are two problems
with these measures.  First, they are two numbers, neither useful
without the other.  Second, they are subject to considerable noise.
In pilot experiments, we found that as we changed our thresholding
values monotonically, precision and recall changed non-monotonically
(see Figure \ref{fig:smoothit}).  We attribute this to the fact that
we must choose a single parse from our parse forest, and, as we
tighten a thresholding parameter, we may threshold out either good or
bad parses.  Furthermore, rather than just changing precision or
recall by a small amount, a single thresholded item may completely
change the shape of the resulting tree.  Thus, precision and recall
are only smooth with very large sets of test data.  However, because
of the large number of experiments we wished to run, using a large set
of test data was not feasible.  Thus, we looked for a surrogate
measure, and decided to use the total inside probability of all
parses, which, with no thresholding, is just the probability of the
sentence given the model.  If we denote the total inside probability
with no thresholding by $I$ and the total inside probability with
thresholding by $I_T$, then $\frac{I_T}{I}$ is the probability that we
did not threshold out the correct parse, given the model.  Thus,
maximizing $I_T$ should maximize correctness.  Since probabilities can
become very small, we instead minimize entropies, the negative
logarithm of the probabilities.  Figure \ref{fig:smoothit} shows that
with a large data set, entropy correlates well with precision and
recall, and that with smaller sets, it is much smoother.  Entropy is
smoother because it is a function of many more variables: in one
experiment, there were about 16000 constituents which contributed to
precision and recall measurements, versus 151 million productions
potentially contributing to entropy.  Thus, we choose entropy as our
measure of correctness for most experiments.  When we did measure
precision and recall, we used the metric as defined by
\newcite{Collins:96a}.

Note that the fact that entropy changes smoothly and monotonically is
critical for the performance of the multiple parameter optimization
algorithm.  Furthermore, we may have to run quite a few iterations of
that algorithm to get convergence, so the fact that entropy is smooth
for relatively small numbers of sentences is a large help.  Thus, the
discovery that entropy is a good surrogate for precision and recall is
non-trivial.  The same kinds of observations could be extended to
speech recognition to optimize multiple thresholds there (the typical
modern speech system has quite a few thresholds), a topic for future
research.  

Note that for some sentences, with too tight thresholding, the parser
will fail to find any parse at all.  We dealt with these cases by
restarting the parser with all thresholds lowered by a factor of 5,
iterating this loosening until a parse could be found.  This is why
for some tight thresholds, the parser may be slower than with looser
thresholds: the sentence has to be parsed twice, once with tight
thresholds, and once with loose ones.

\begin{figure}[htb]
\psfig{figure=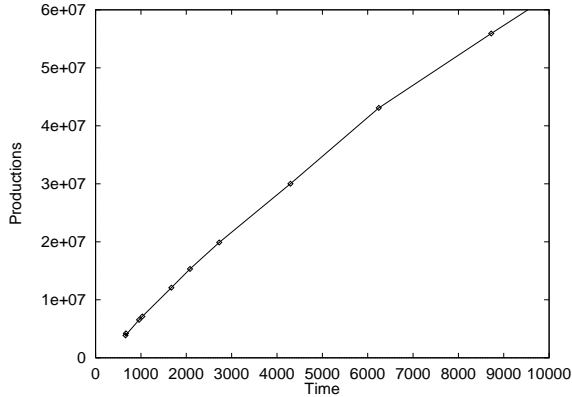,width=3in}
\caption{Productions versus Time} \label{fig:timevsprod}
\end{figure}

Next, we needed to choose a measure of time.  There are two obvious
measures: amount of work done by the parser, and elapsed time.  If we
measure amount of work done by the parser in terms of the number of
productions with non-zero probability examined by the parser, we have
a fairly implementation-independent, machine-independent measure of
speed.  On the other hand, because we used many different thresholding
algorithms, some with a fair amount of overhead, this measure seems
inappropriate.  Multiple-pass parsing requires use of the outside
algorithm; global thresholding uses its own dynamic programming
algorithm; and even beam thresholding has some per-node overhead.
Thus, we will give most measurements in terms of elapsed time, not
including loading the grammar and other $O(1)$ overhead.  We did want to
verify that elapsed time was a reasonable measure, so we did a beam
thresholding experiment to make sure that elapsed time and number of
productions examined were well correlated, using 200 sentences and an
exponential sweep of the thresholding parameter.  The results, shown
in Figure \ref{fig:timevsprod}, clearly indicate that time is a good
proxy for productions examined.

\subsection{Experiments in Beam Thresholding}

\label{sec:beamexp}

\begin{figure*}
\begin{center}
\begin{tabular}{cc}
Precision/Recall & Total Entropy \\
15 Sentences & 15 Sentences \\
\psfig{figure=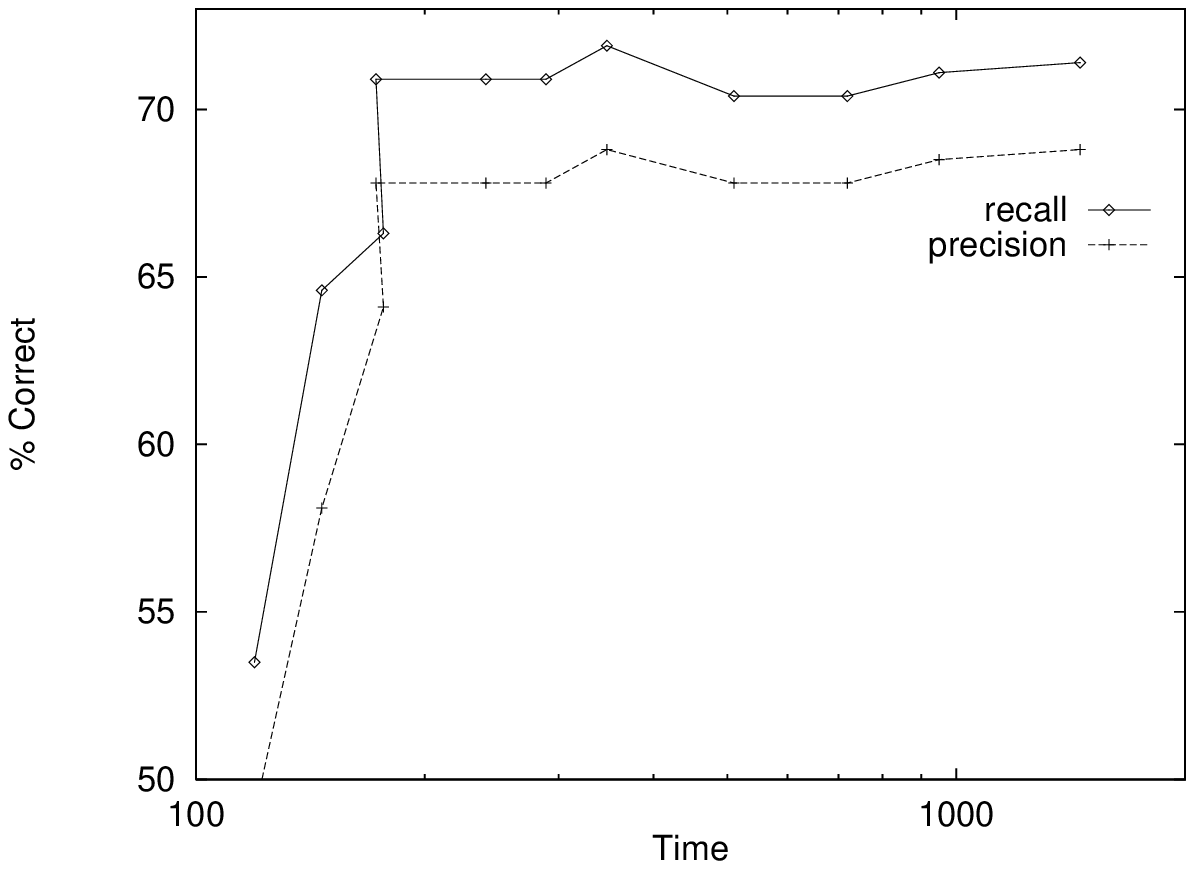,width=3.1in} &
\psfig{figure=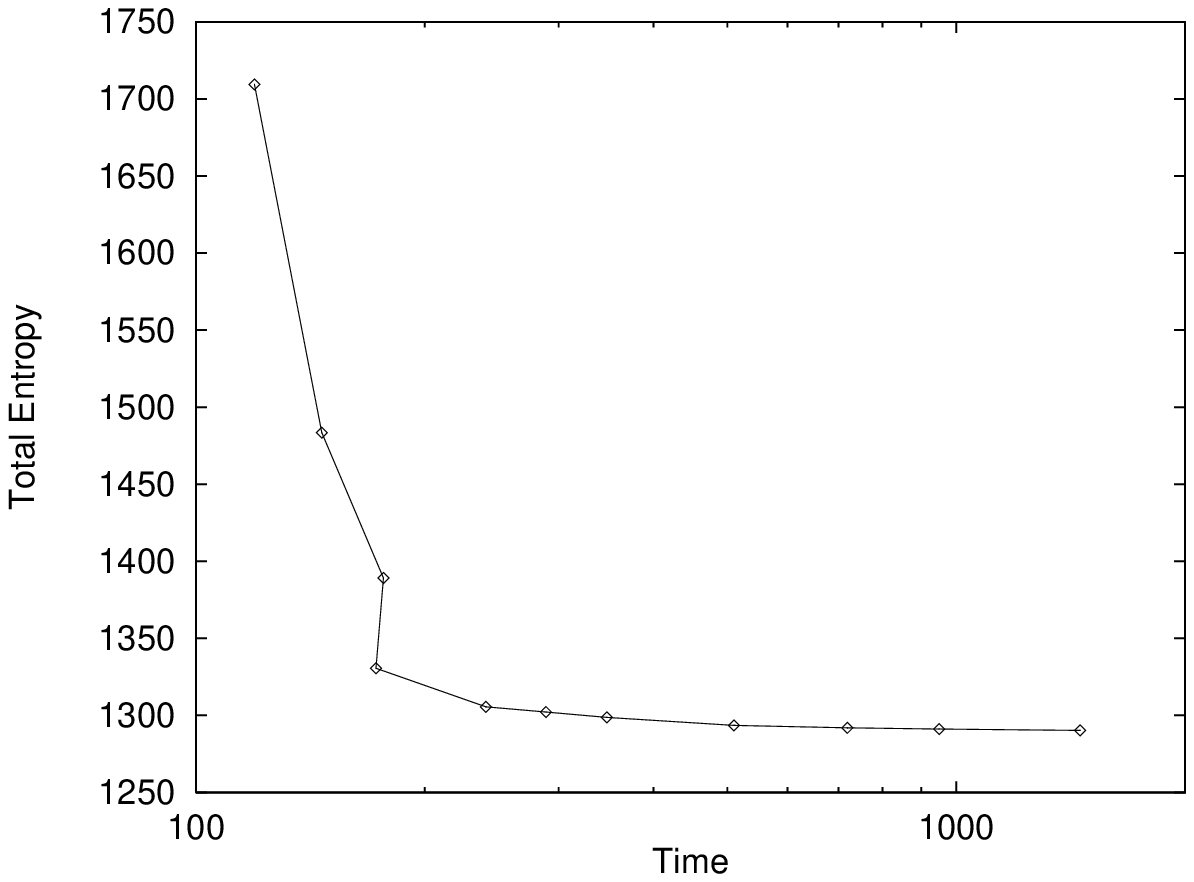,width=3.1in} \\
Precision/Recall & Total Entropy \\
200 Sentences & 200 Sentences \\
\psfig{figure=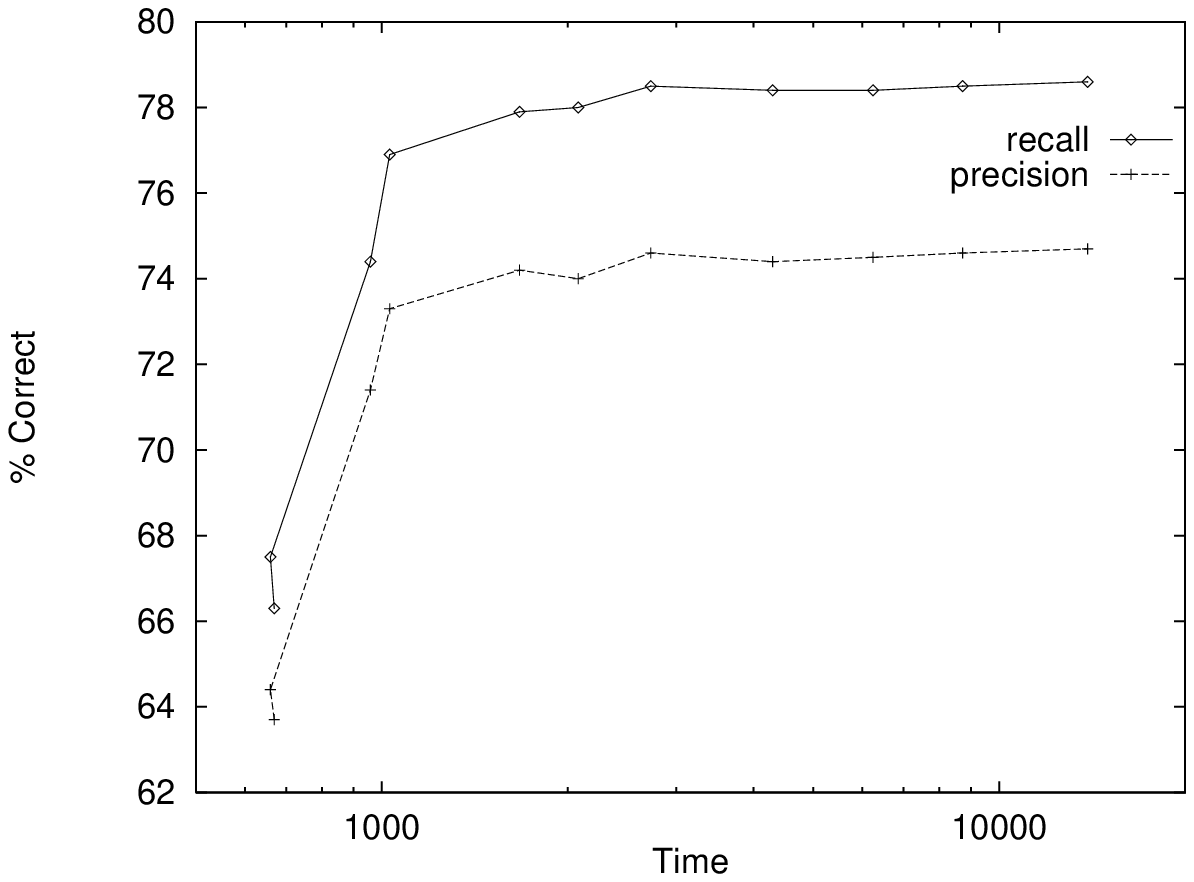,width=3.1in} &
\psfig{figure=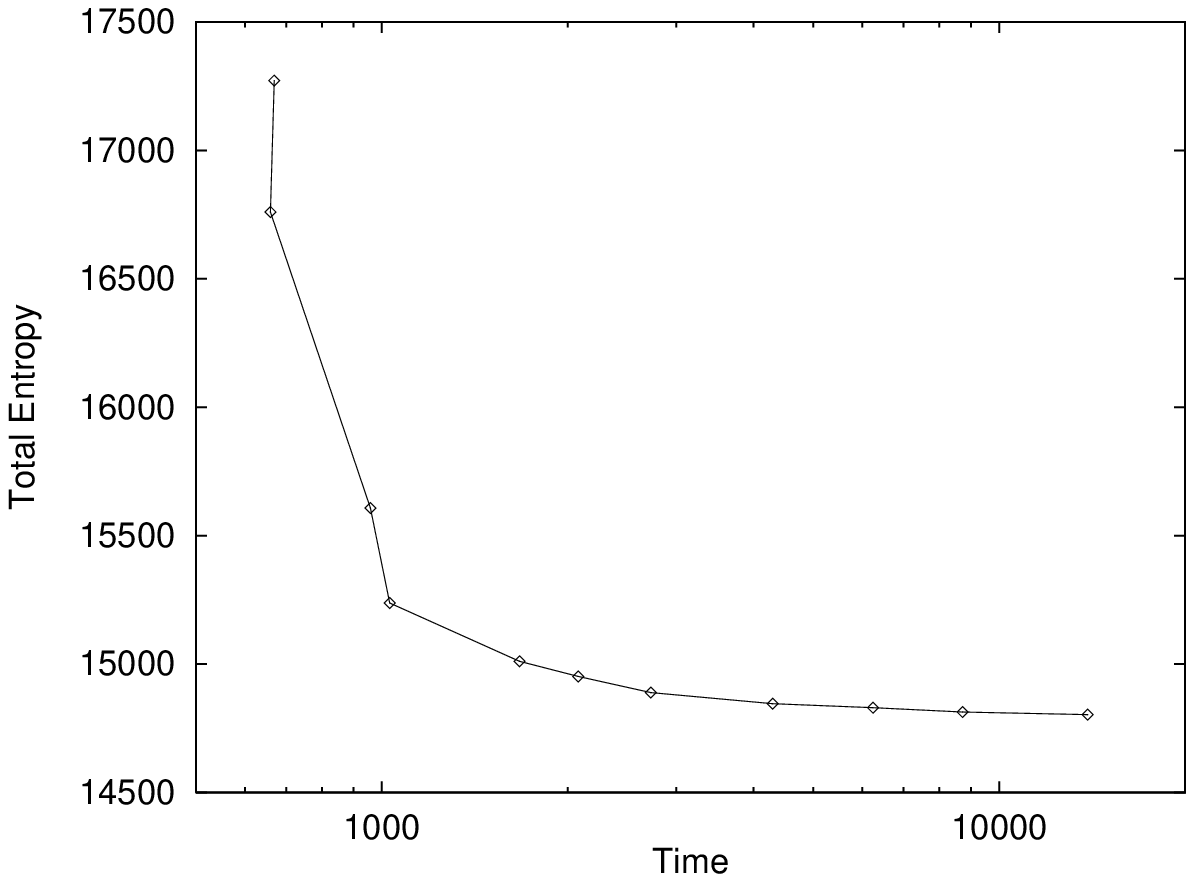,width=3.1in} \\
\end{tabular}
\caption{Smoothness for Precision and Recall versus Total Inside for
Different Test Data Sizes} \label{fig:smoothit}
\end{center}
\end{figure*}

\begin{figure*}
\begin{tabular}{cc}
\psfig{figure=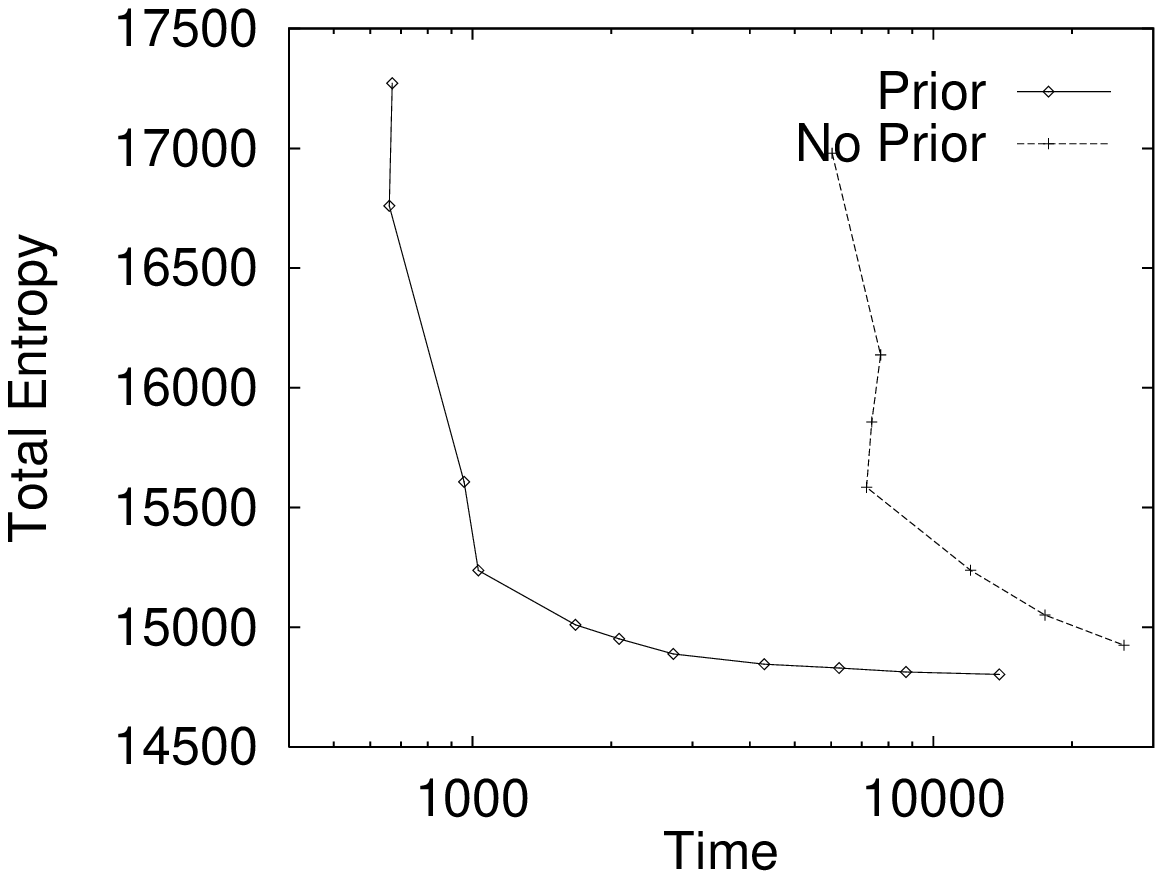,width=3.1in} &
\psfig{figure=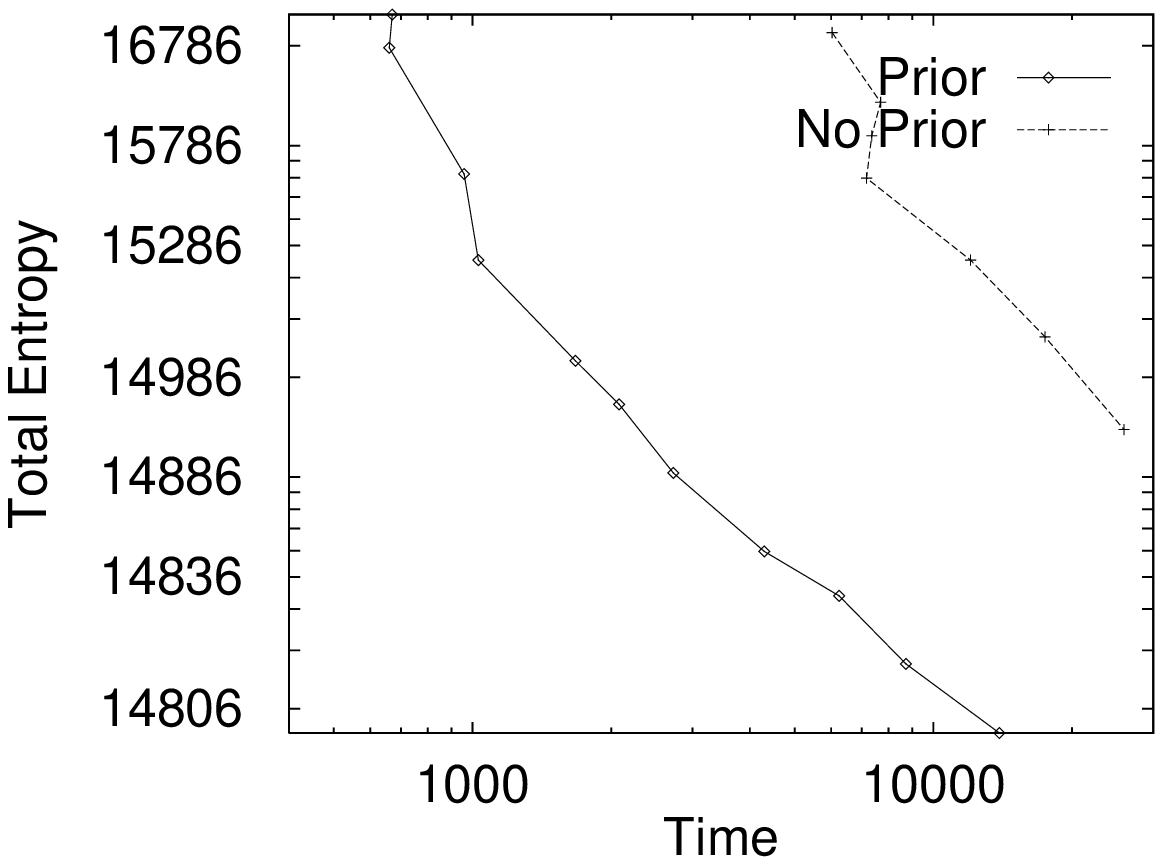,width=3.1in} \\ 
\hspace{.3in}$X$ axis: $\mathit{log(time)}$ &
\hspace{.3in}$X$ axis: $\mathit{log(time)}$ \\
\hspace{.3in}$Y$ axis: $\mathit{entropy}$ &
\hspace{.3in}$Y$ axis: $\mathit{log}(\mathit{entropy}-\mathit{asymptote})$
\\
\end{tabular}
\caption{Beam Thresholding with and without the Prior Probability,
Two Different Scales}
\label{fig:priorvsno}
\end{figure*}

Our first goal was to show that entropy is a good surrogate for
precision and recall.  We thus tried two experiments: one with a
relatively large test set of 200 sentences, and one with a relatively
small test set of 15 sentences.  Presumably, the 200 sentence test set
should be much less noisy, and fairly indicative of performance.  We
graphed both precision and recall, and entropy, versus time, as we
swept the thresholding parameter over a sequence of values.  The
results are in Figure \ref{fig:smoothit}.  As can be seen, entropy is
significantly smoother than precision and recall for both size test
corpora.

Our second goal was to check that the prior probability is indeed
helpful.  We ran two experiments, one with the prior and one without.
Since the experiments without the prior were much worse than those
with it, all other beam thresholding experiments included the prior.
The results, shown in Figure \ref{fig:priorvsno}, indicate that the
prior is a critical component.  This experiment was run on 200
sentences of test data.

Notice that as the time increases, the data tends to approach an
asymptote, as shown in the left hand graph of Figure
\ref{fig:priorvsno}.  In order to make these small asymptotic changes
more clear, we wished to expand the scale towards the asymptote.  The
right hand graph was plotted with this expanded scale, based on
$\mathit{log}(\mathit{entropy}-\mathit{asymptote})$, a slight
variation on a normal log scale.  We use this scale in all the
remaining entropy graphs.  A normal logarithmic scale is used for the
time axis.  The fact that the time axis is logarithmic is especially
useful for determining how much more efficient one algorithm is than
another at a given performance level.  If one picks a performance
level on the vertical axis, then the distance between the two curves
at that level represents the ratio between their speeds.  There is
roughly a factor of 8 to 10 difference between using the prior and not
using it at all graphed performance levels, with a slow trend towards
smaller differences as the thresholds are loosened.

\subsection{Experiments in Global Thresholding}

We tried experiments comparing global thresholding to beam
thresholding.  Figure \ref{fig:beamjob} shows the results of this
experiment, and later experiments.  In the best case, global
thresholding works twice as well as beam thresholding, in the sense
that to achieve the same level of performance requires only half as
much time, although smaller improvements were more typical.

We have found that, in general, global thresholding works better on
simpler grammars.  In some complicated grammars we explored in other
work, there were systematic, strong correlations between nodes, which
violated the independence approximation used in global thresholding.
This prevented us from using global thresholding with these grammars.
In the future, we may modify global thresholding to model some of
these correlations.  

\subsection{Experiments combining Global Thresholding and Beam Thresholding}

\begin{figure*}
\psfig{figure=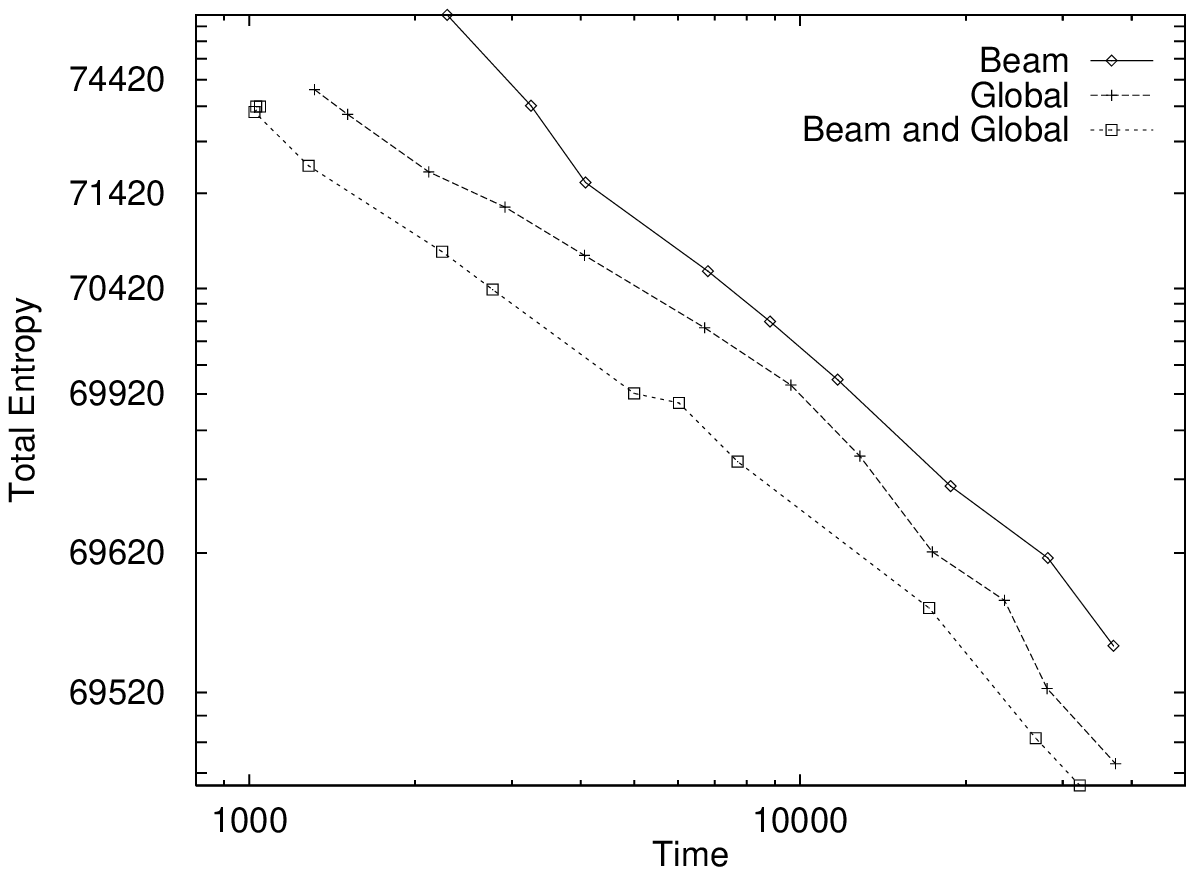,width=6in} 
\caption{Combining Beam and Global Search}
\label{fig:beamjob}
\end{figure*}

While global thresholding works better than beam thresholding in
general, each has its own strengths.  Global thresholding can
threshold across cells, but because of the approximations used, the
thresholds must generally be looser.  Beam thresholding can only
threshold within a cell, but can do so fairly tightly.  Combining the
two offers the potential to get the advantages of both.  We ran a
series of experiments using the thresholding optimization algorithm of
Section \ref{sec:descent}.  Figure \ref{fig:beamjob} gives the
results.  The combination of beam and global thresholding together is
clearly better than either alone, in some cases running 40\% faster
than global thresholding alone, while achieving the same performance
level.  The combination generally runs twice as fast as beam
thresholding alone, although up to a factor of three.

\subsection{Experiments in Multiple-Pass Parsing}

\label{sec:expmulti}

Multiple-pass parsing improves even further on our experiments
combining beam and global thresholding.  Note that we used both beam
and global thresholding for both the first and second pass in these
experiments.  The first pass grammar was the very simple
terminal-prime grammar, and the second pass grammar was the usual
6-gram grammar.

We evaluated multiple-pass parsing slightly differently from the other
thresholding techniques.  In the experiments conducted here, our first
and second pass grammars were very different from each other.  For a
given parse to be returned, it must be in the intersection of both
grammars, and reasonably likely according to both.  Since the first
and second pass grammars capture different information, parses which
are likely according to both are especially good.  The entropy of a
sentence measures its likelihood according to the second pass, but
ignores the fact that the returned parse must also be likely according
to the first pass.  Thus, entropy, our measure in the previous
experiments, which measures only likelihood according to the final
pass, is not necessarily the right measure to use.  We therefore give
precision and recall results in this section.  We still optimized our
thresholding parameters using the same 31 sentence held out corpus,
and minimizing entropy versus number of productions, as before.

\begin{figure*}
\psfig{figure=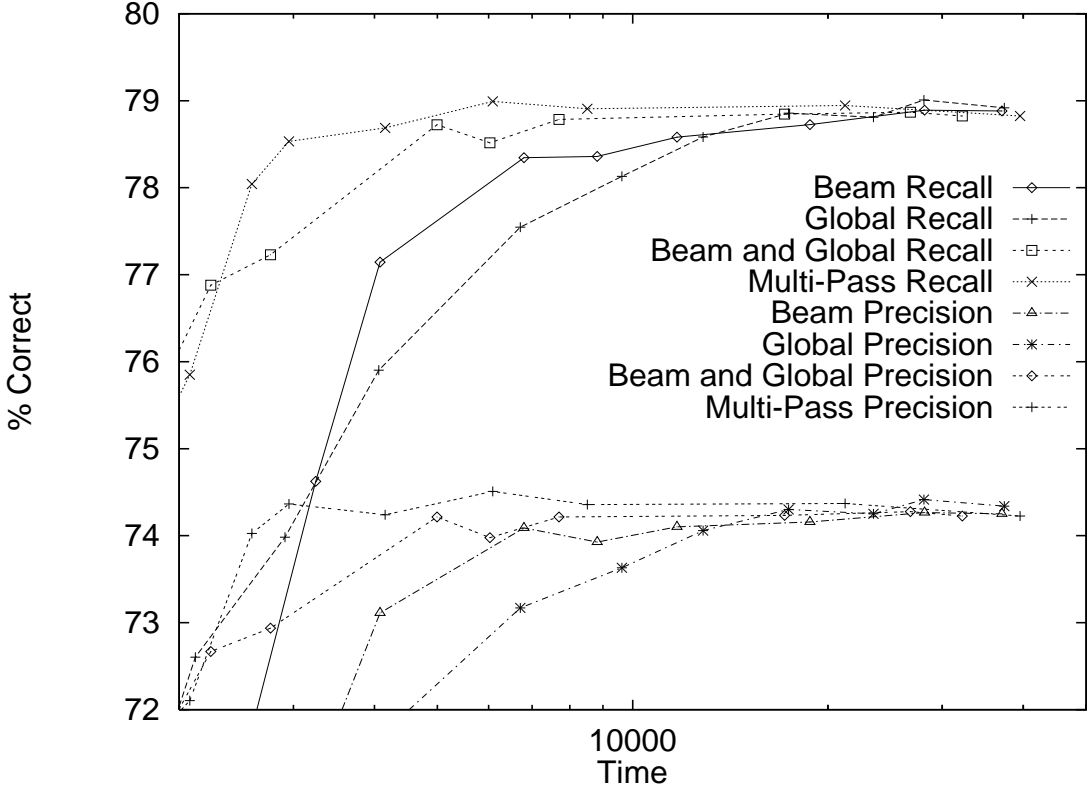,width=6in} 
\caption{Multiple Pass Parsing vs. Beam and Global vs. Beam}
\label{fig:multiexp}
\end{figure*}

We should note that when we used a first pass grammar that captured a
strict subset of the information in the second pass grammar, we have
found that entropy is a very good measure of performance.  As in our
earlier experiments, it tends to be well correlated with precision and
recall but less subject to noise.  It is only because of the grammar
mismatch that we have changed the evaluation.

Figure \ref{fig:multiexp} shows precision and recall curves for single
pass versus multiple pass experiments.  As in the entropy curves, we
can determine the performance ratio by looking across horizontally.
For instance, the multi-pass recognizer achieves a 74\% recall level
using 2500 seconds, while the best single pass algorithm requires
about 4500 seconds to reach that level.  Due to the noise resulting
from precision and recall measurements, it is hard to exactly quantify
the advantage from multiple pass parsing, but it is generally about
50\%.

\section{Applications and Conclusions}

\subsection{Application to Other Formalisms}

In this paper, we only considered applying multiple-pass and global
thresholding techniques to parsing probabilistic context-free
grammars.  However, just about any probabilistic grammar formalism for
which inside and outside probabilities can be computed can benefit
from these techniques.  For instance, Probabilistic Link Grammars
\cite{Lafferty:92a} could benefit from our algorithms.  We have
however had trouble using global thresholding with grammars that
strongly violated the independence assumptions of global thresholding.

One especially interesting possibility is to apply multiple-pass
techniques to formalisms that require $\gg O(n^3)$ parsing time, such
as Stochastic Bracketing Transduction Grammar (SBTG) \cite{Wu:96a} and
Stochastic Tree Adjoining Grammars (STAG)
\cite{Resnik:92a,Schabes:92a}.  SBTG is a context-free-like formalism
designed for translation from one language to another; it uses a four
dimensional chart to index spans in both the source and target
language simultaneously.  It would be interesting to try speeding up
an SBTG parser by running an $O(n^3)$ first pass on the source language
alone, and using this to prune parsing of the full SBTG.

The STAG formalism is a mildly context-sensitive formalism, requiring
$O(n^6)$ time to parse.  Most STAG productions in practical grammars
are actually context-free.  The traditional way to speed up STAG
parsing is to use the context-free subset of an STAG to form a
Stochastic Tree Insertion Grammar (STIG) \cite{Schabes:94a}, an
$O(n^3)$ formalism, but this method has problems, because the STIG
undergenerates since it is missing some elementary trees.  A different
approach would be to use multiple-pass parsing.  We could first find a
context-free covering grammar for the STAG, and use this as a first
pass, and then use the full STAG for the second pass.

\subsection{Conclusions}

The grammars described here are fairly simple, presented for purposes
of explication.  In other work in preparation, in which we have used a
significantly more complicated grammar, which we call the
Probabilistic Feature Grammar (PFG), the
improvements from multiple-pass parsing are even more dramatic: single
pass experiments are simply too slow to run at all.

We have also found the automatic thresholding parameter optimization
algorithm to be very useful.  Before writing the parameter
optimization algorithm, we developed the PFG grammar and the
multiple-pass parsing technique and ran a series of experiments using
hand optimized parameters.  We recently ran the optimization algorithm
and reran the experiments, achieving a factor of two speedup with no
performance loss.  While we had not spent a great deal of time hand
optimizing these parameters, we are very encouraged by the
optimization algorithm's practical utility.

This paper introduces four new techniques: beam thresholding with
priors, global thresholding, multiple-pass parsing, and automatic
search for thresholding parameters.  Beam thresholding with priors can
lead to almost an order of magnitude improvement over beam
thresholding without priors.  Global thresholding can be up to three
times as efficient as the new beam thresholding technique, although
the typical improvement is closer to 50\%.  When global thresholding
and beam thresholding are combined, they are usually two to three
times as fast as beam thresholding alone.  Multiple-pass parsing can
lead to up to an additional 50\% improvement with
the grammars in this paper.  We expect the parameter optimization
algorithm to be broadly useful.


\bibliography{master}

\end {document}